\documentclass[10pt,twocolumn,letterpaper]{article}

\usepackage{cvpr}              %
\usepackage{graphicx}
\usepackage{amsmath}
\usepackage{amssymb}
\usepackage{booktabs}
\usepackage{caption}
\pdfoutput=1
\usepackage{multirow}
\usepackage{lipsum}
\usepackage{longtable}
\usepackage{svg}
\usepackage{float}
\usepackage{array}
\usepackage[accsupp]{axessibility}  %
\newcolumntype{P}[1]{>{\centering\arraybackslash}p{#1}}
\newcolumntype{M}[1]{>{\centering\arraybackslash}m{#1}}
\usepackage{algorithm}
\usepackage[margin=1in]{geometry}
\usepackage{xcolor}  %
\usepackage{algpseudocode}
\usepackage{rotating}
\usepackage{adjustbox}
\usepackage[pagebackref,breaklinks,colorlinks,bookmarks=false]{hyperref}
\usepackage{subcaption}

\usepackage[capitalize]{cleveref}
\crefname{section}{Sec.}{Secs.}
\Crefname{section}{Section}{Sections}
\Crefname{table}{Table}{Tables}
\crefname{table}{Tab.}{Tabs.}

\def\cvprPaperID{24} %
\def\confName{CVPR}
\def\confYear{2025}

\newcommand*{\affaddr}[1]{#1} %
\newcommand*{\affmark}[1][*]{\textsuperscript{#1}}
\newcommand*{\email}[1]{\texttt{#1}}
\begin{document}

\title{Seeing like a Cephalopod: Colour Vision with a Monochrome Event Camera}

\author{%
Sami Arja\affmark[*], Nimrod Kruger, Alexandre Marcireau, Nicholas Owen Ralph \\ Saeed Afshar and Gregory Cohen\\
\rule{0pt}{3ex} \affaddr{Western Sydney University}\\
\email{$^{*}$s.elarja@westernsydney.edu.au}\\
}

\thispagestyle{empty}
\twocolumn[{
    \renewcommand\twocolumn[1][]{#1}
    \maketitle
    \includegraphics[width=\textwidth]{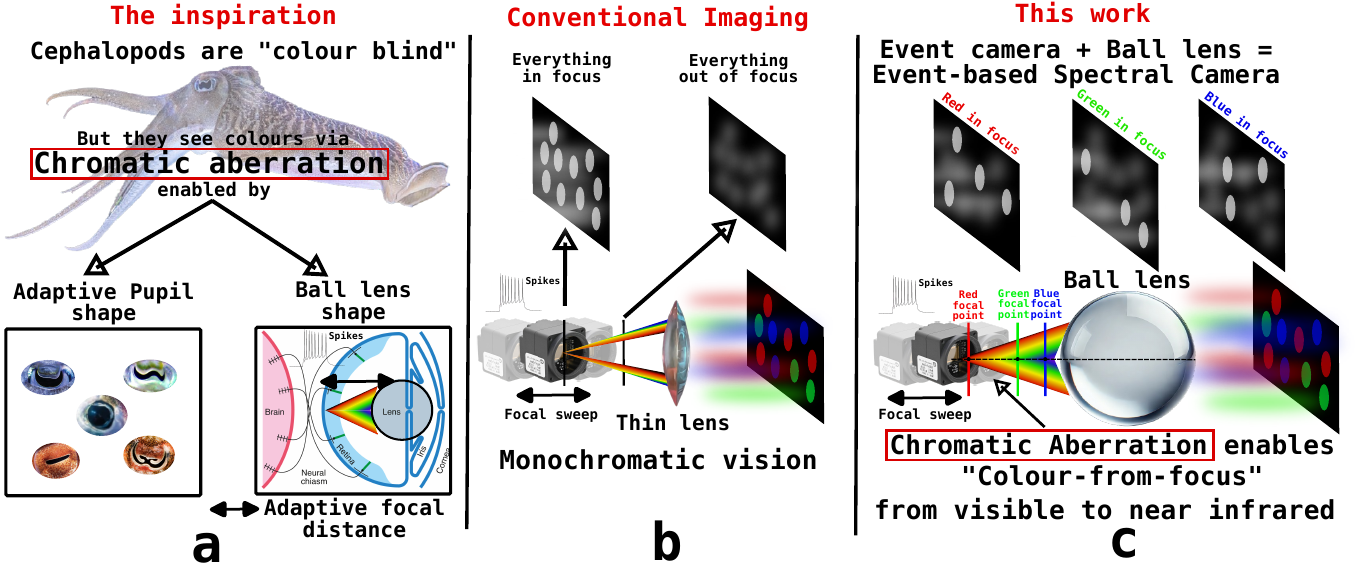}
    \captionof{figure}{The Neuromorphic Cephalopod Eye. \textbf{(a) The inspiration:} Cephalopods are hypothesised to perceive colours, despite having only one photoreceptor type (monochromatic vision), by exploiting chromatic aberration from their ball-shaped lens and uniquely shaped pupil. The lens refracts wavelengths differently, causing distinct focal lengths for each wavelength, while the irregular pupil maximises chromatic blur. This transforms spectral information from the environment into spatial patterns on the retina, enabling colour perception. \textbf{(b) Conventional imaging:} Conventional event cameras are mostly monochromatic and use standard lenses to focus; the scene can either be in focus or out of focus, regardless of the spectral information in the scene. \textbf{(c) This work:} Inspired by the cephalopod's ability to perceive spectral information through chromatic aberration and by focusing, this paper shows that it is possible to convert a monochromatic event camera into a spectral imaging sensor capable of distinguishing colours from visible to near-infrared by using a ball lens controlled by a linear actuator.} \vspace{2em}
    \label{fig:long}
}]

\begin{abstract}

   Cephalopods exhibit unique colour discrimination capabilities despite having one type of photoreceptor, relying instead on chromatic aberration induced by their ocular optics and pupil shapes to perceive spectral information. We took inspiration from this biological mechanism to design a spectral imaging system that combines a ball lens with an event-based camera. Our approach relies on a motorised system that shifts the focal position, mirroring the adaptive lens motion in cephalopods. This approach has enabled us to achieve wavelength-dependent focusing across the visible light and near-infrared spectrum, making the event a spectral sensor. We characterise chromatic aberration effects, using both event-based and conventional frame-based sensors, validating the effectiveness of bio-inspired spectral discrimination both in simulation and in a real setup, as well as assessing the spectral discrimination performance. Our proposed approach provides a robust spectral sensing capability without conventional colour filters or computational demosaicing. This approach opens new pathways toward new spectral sensing systems inspired by nature's evolutionary solutions. Code and analysis are available at: \href{https://samiarja.github.io/neuromorphic_octopus_eye/}{Project Page}.
\end{abstract}

\section{Introduction}
\label{sec:intro}
Cephalopods such as octopus, squid, and cuttlefish can perform colour camouflage and signalling despite being "colour-blind" by conventional definition~\cite{messenger_evidence_1973}. They possess only a single type of pigment in their photoreceptors, yet they can match and display a wide range of colours in their environment. Previous works~\cite{stubbs_spectral_2016,jagger_wide-angle_1999} uncovered an optical mechanism for spectral discrimination that does not require multiple receptor types. In essence, cephalopod eyes exploit a chromatically aberrated imaging system in combination with uniquely shaped pupils to infer colour from focus. Cephalopods' vision is comprised of a ball-shaped lens for focusing light on the retina, different wavelengths come into focus at different distances behind the lens, which is achieved by moving the lens core relative to the retina, which enables projecting various focal planes and determining which setting yields the sharpest contrast~\cite{stubbs_spectral_2016}. This chromatic blur method provides coarse spectral information: a scan through the various focal distances effectively scans the scene’s predominant colour, which has been suggested to allow the animal to deduce colour by finding the best focus for maximum contrast~\cite{stubbs_spectral_2016}. A distinctive property of cephalopod ball lenses is their graded refractive index, where the refractive index is highest at the lens core and decreases radially outward, approaching that of the surrounding medium at the outer surface~\cite{jagger_wide-angle_1999,cai_eye_2017}. A graded index (GRIN) feature can explain how spherical aberrations are corrected when focusing an object on the retina with the correct retina-lens distance, but contributes little to correcting chromatic aberrations ~\cite{jagger_wide-angle_1999}.

Many shallow-water cephalopods have evolved off-axis or complex pupil shapes (e.g., U- or W-shaped pupils). The combination of these specialised pupils and ball lenses projects distinct spatial patterns onto the retina, depending on the spectral content and position of incoming light. Elongated or semi-annular pupils collect light at oblique angles, intentionally enhancing chromatic blur and spectral cues, though at the expense of spatial sharpness~\cite{stubbs_spectral_2016}, prioritising spectral discrimination over spatial resolution.

Compared to organisms with multi-receptor colour vision, which divide light into separate channels, reducing sensitivity and resolution, cephalopods may achieve a form of colour vision by using a single broadband receptor with optical tuning. Their large optic lobes likely process the focus-dependent contrast information to resolve colour, compensating for the more complex computational task. This unique strategy of colour perception via chromatic aberration and pupil design was the main inspiration for this work.

\begin{figure}[ht]
  \centering
  \includegraphics[width=3.1in]{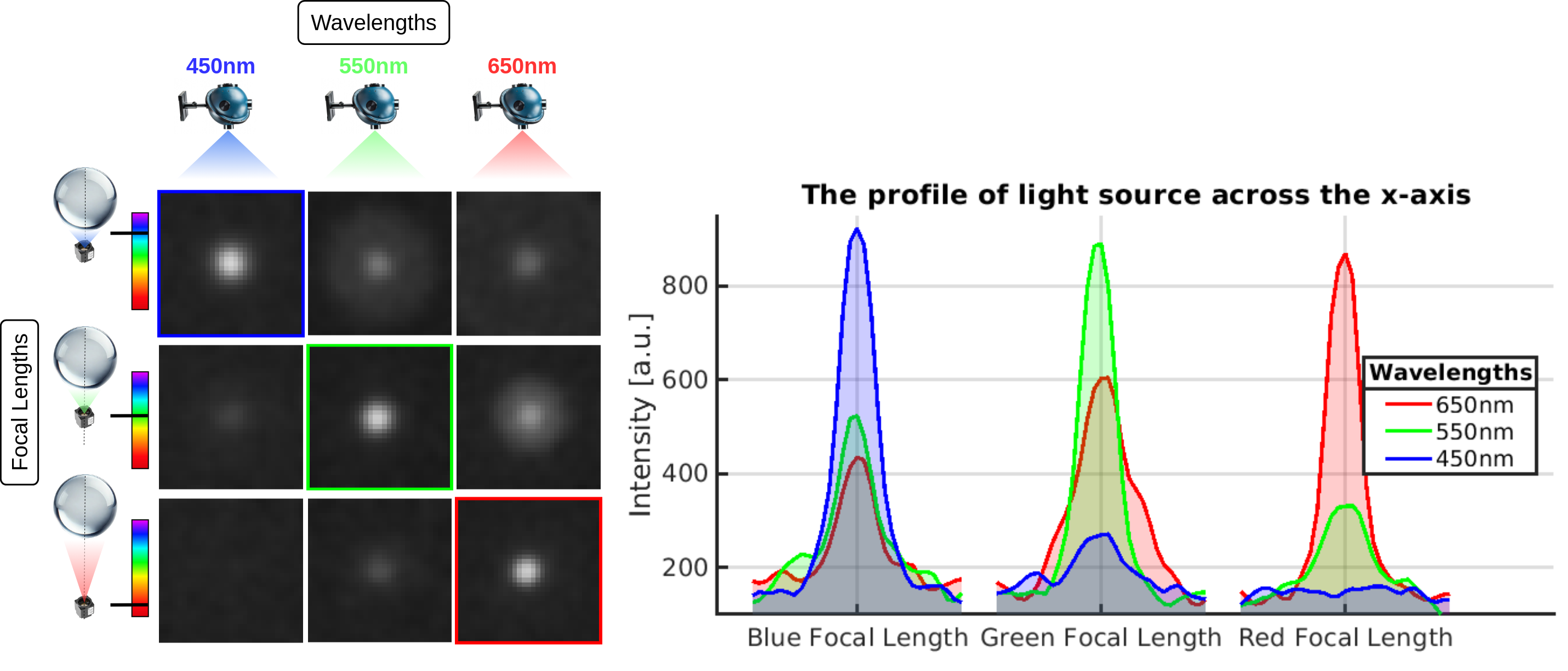}
  \caption{Effects of chromatic aberration. \textbf{Left:} Demonstration of selective focusing of different wavelengths by adjusting the focal distance, captured using a Sony IMX249 frame camera sensor. \textbf{Right:} Horizontal cross-section of the image at each focal distance, illustrating that each wavelength has a unique optimal focal plane.} 
  \label{fig:thinball_lens_concept}
\end{figure}

Chromatic aberration is illustrated in Figure~\ref{fig:long}(c). Unlike an ideal thin lens, which focuses all wavelengths at the same point, resulting in a uniformly sharp or blurry image (Figure~\ref{fig:long}(b)), a ball lens introduces both spherical and chromatic aberrations. This causes different wavelengths to focus at distinct distances. This focal shift degrades a white point object into a series of superimposed unfocused coloured discs. By precisely adjusting the imager's position relative to the lens, each wavelength can individually be brought into sharp focus and defocus the other wavelengths, as demonstrated in Figure~\ref{fig:thinball_lens_concept}, by showing how the Point Spread Functions (PSFs) for blue (450 nm), green (550 nm), and red (650 nm) wavelengths can focus at their respective focal lengths. Each PSF is sharp only at its specific focal point and quickly becomes blurred as the imaging plane moves away.

Our contributions in this paper are summarised as follows:

\begin{itemize}
    \item \textbf{Event-based Colour from Focus:} A demonstration of a cephalopod-inspired imaging technique that enables an event camera to perceive spectral information in visible light and near-infrared by exploiting chromatic aberration from a ball lens.
    \item \textbf{Simulation of Imaging Technique:} The development of a computational simulator and optical experimental setup to validate and test the spectral discrimination and assess the contribution of chromatic aberration of the system with respect to focal distance. Section~\ref{sec:realsetup} and Section~\ref{sec:simulator}.
    \item \textbf{Spectral Characterisation and Analysis:} Investigation of the spectral discrimination performance across a wide spectrum to provide a deep understanding of the sensor response in frames and events domains. Section~\ref{sec:Results}.
\end{itemize}

\subsection{Motivation}
\label{sec:motiv}

Understanding and taking inspiration from biological systems has long inspired innovations in engineering, particularly in vision systems, notably the event camera~\cite{lichtsteiner_128times128_2008}. Traditional approaches for capturing colour information in event-based imaging systems typically involve Bayer filters or discrete spectral filters~\cite{berner_self-timed_2008, lenero-bardallo_bio-inspired_2014, farian_bio-inspired_2015,li_design_2015, taverni_front_2018,marcireau_event-based_2018}. While effective, colour information cannot be directly inferred from the events unless reconstructed or with the use of exposure frames. Recent works have offered intriguing insights into how organisms with inherently monochromatic vision, such as cephalopods, may achieve colour discrimination~\cite{stubbs_spectral_2016}, enabled by chromatic aberration produced by the ball lenses in their eyes. Cephalopods, despite having only a single type of photoreceptor, have a sophisticated neural mechanism involving lateral inhibition to perceive colour through spike-based visual signals~\cite{nahmad-rohen_contrast_2019}.

Motivated by these biological concepts, our work explores an alternative approach to colour vision, inspired directly by the visual system of cephalopods. We leverage chromatic aberration, a phenomenon typically viewed as a drawback in optics, as an advantageous mechanism to see colours. By using a ball lens, we demonstrate a simple motion-dependent technique for extracting spectral information without complex filters or extensive post-processing. Although event cameras do not exactly replicate the anatomy of the cephalopod eye, they share critical functional characteristics, including monochromatic sensing and the generation of asynchronous spikes in response to brightness changes~\cite{nahmad-rohen_contrast_2019}.

Our goal is not to replace conventional colour-filter-based methods but rather to present a complementary, biologically inspired solution. Through this approach, we showcase how an optical mechanism enables monochromatic vision sensors with spectral imaging capabilities across VIS and IR wavelengths.

\subsection{Related Work}\label{sec:relatedwork}

\textbf{Optics of Ball lenses}. Ball or spherical lenses have distinctive optical properties that make them both intriguing and challenging for imaging applications. Their extremely short focal lengths relative to their diameter, corresponding to a very high Numerical Aperture (NA), enable the capture and focusing of a large illumination influx, while their very wide angular range enables them to produce ultra-wide-angle images. For a given lens size, a spherical lens achieves one of the smallest focal points, which is why ball lenses are commonly used to couple light into optical fibers and in micro-optical systems where maximising light collection is critical~\footnote{\url{https://www.edmundoptics.com.au/knowledge-center/application-notes/optics/understanding-ball-lenses/?srsltid=AfmBOopJfnTeI9fA9FAfjzzDKkfizDBLJDS8w0_-W-9i1U-o8YqjdehM}}. The symmetry of a ball lens also gives it an omnidirectional field of view.

However, these advantages come at the cost of significant spherical and chromatic aberrations. Spherical aberrations are introduced when rays entering near the lens edges focus at different points than those near the centre, leading to a blurry image. Chromatic aberration (like any uncorrected lens) will focus different wavelengths (colours) at different distances due to the dispersion of the glass. In practical terms, a ball lens used in imaging will produce colour fringing and defocus for off-nominal wavelengths or broad-spectrum illumination.

A ball lens has a wavelength-dependent refraction index $n_{\text{lens}}$, causing significant spherical and chromatic aberrations and producing unique focal points for each wavelength $\lambda$. To determine how the ball lens focuses light along the optical axis, the ball lens (with a diameter $D$) is typically described by its Effective Focal Length (EFL), measured from the lens centre, and Back Focal Length (BFL) measured from the rear surface:\small
\begin{equation}
\mathrm{EFL}(\lambda)=\frac{n_{\mathrm{lens}}(\lambda)D}{4\bigl(n_{\mathrm{lens}}(\lambda)-1\bigr)}\quad,\quad\mathrm{BFL}(\lambda)=\mathrm{EFL}(\lambda)-\frac{D}{2},
\end{equation}
Due to the nonlinearity of the refractive index $n_{\mathrm{lens}}(\lambda)$ for each wavelength, the induced chromatic aberration shifts the focal plane for different wavelengths non-linearly. For example, the distance between focal planes for two adjacent "blue" wavelengths (e.g. 400 \& 450 nm) is larger than for two adjacent "red" wavelengths (e.g. 650 \& 700 nm).

The lens equation and magnification rate of the ball lens for an object at distance $d_o$ and image at $d_{i}(\lambda)$ are described as,
\begin{equation}
\frac{1}{\mathrm{EFL}(\lambda)}=\frac{1}{d_o}+\frac{1}{d_i(\lambda)},\quad\text{and}\quad M(\lambda)=-\frac{d_i(\lambda)}{d_o},\end{equation}
leading to different focal planes and magnifications depending on the source position and spectral composition.

To our knowledge, this is the first work that uses a ball lens on an event camera to extract spectral information from the chromatic aberration produced by the lens.

\textbf{Event-based colour Perception}. Various methods have emerged for colour sensing in neuromorphic imaging systems. Initial approaches involved vertically stacked photodiodes, providing high-resolution colour information by capturing multiple wavelengths at the same spatial location \cite{berner_self-timed_2008, lenero-bardallo_bio-inspired_2014, farian_bio-inspired_2015}. An alternative method employs Bayer filtering, or Colour Filter Arrays (CFA), where individual pixels are overlaid with discrete colour filters, effectively reducing spatial resolution but simplifying the sensor architecture \cite{li_design_2015, taverni_front_2018}. Additionally, a three-chip method combines three separate event cameras, each equipped with distinct colour filters, and uses beam splitters to separate incoming light into three spectral bands. This approach maintains full spatial resolution across each wavelength channel~\cite{shimonomura_color_2011, marcireau_event-based_2018}.

\cite{fu_cmos_2009} designed a chip to enable colour differentiation, and \cite{berner_event-based_2011} proposed the first commercially available Colour Dynamic and Active-pixel Vision Sensor (CDAVIS), which has a single buried double-junction photodiode to detect brightness and wavelength changes. However, the CDAVIS approach was limited due to larger pixel sizes and inadequate colour separation. Recent advancements, such as the ColourDAVIS346 sensor \cite{taverni_front_2018}, have improved sensor performance through backside illumination technology and Bayer filters (a 2x2 filter matrix in the order of RGGB placed over the pixel sensors), resulting in higher sensitivity, quantum efficiency, and improved fill factor. The first colour event data was made available by \cite{scheerlinck_ced_2019} from a CDAVIS, and its performance has been further investigated for agricultural~\cite{el2022neuromorphic} and neural imaging applications~\cite{moeys2017sensitive}.

We replaced traditional three-band colour filters, which require complex reconstruction, with a cephalopod-inspired ball lens that captures the entire visible and near-infrared range by adjusting the focal plane. This setup encodes spectral information directly into the event data, without the need for reconstruction algorithms.

\textbf{Active Neuromorphic Sensing}. Our approach builds on the concept of `active neuromorphic sensing', using focal length modulation and optical scanning to improve imaging in event sensors \cite{ralph2023shake} \cite{ralph2024active}. These earlier studies demonstrated that controlled motion scans, inspired by saccadic eye motion, can enhance target detection and spatial resolution \cite{ralph2023shake}. Additionally, variable-focus liquid lenses can be used to create focal length oscillations for better object detection and localisation \cite{ralph2024active}. This method addressed the challenge of rapid and precise focusing in event vision sensors, inspired by biological focus accommodation, where the optimal focus is detected incidentally at each focal length oscillation while simultaneously inducing spatio-temporal contrast. Similar works have recovered depth information using focus variations \cite{martel2017real} \cite{haessig2019spiking} and improved imaging through mutual motion-based contrast \cite{hetexture}, and saccadic motion approaches \cite{oster2007spike} \cite{mishra2017saccade}.  Although these prior works increased temporal contrast to improve general performance, such as target detection, focusing and noise suppression, instead, we have repurposed focal modulation to decode spectral information.

\section{Overview of the Optical Setup}\label{sec:realsetup}

\begin{figure}[h]
  \centering
  \includegraphics[width=\linewidth]{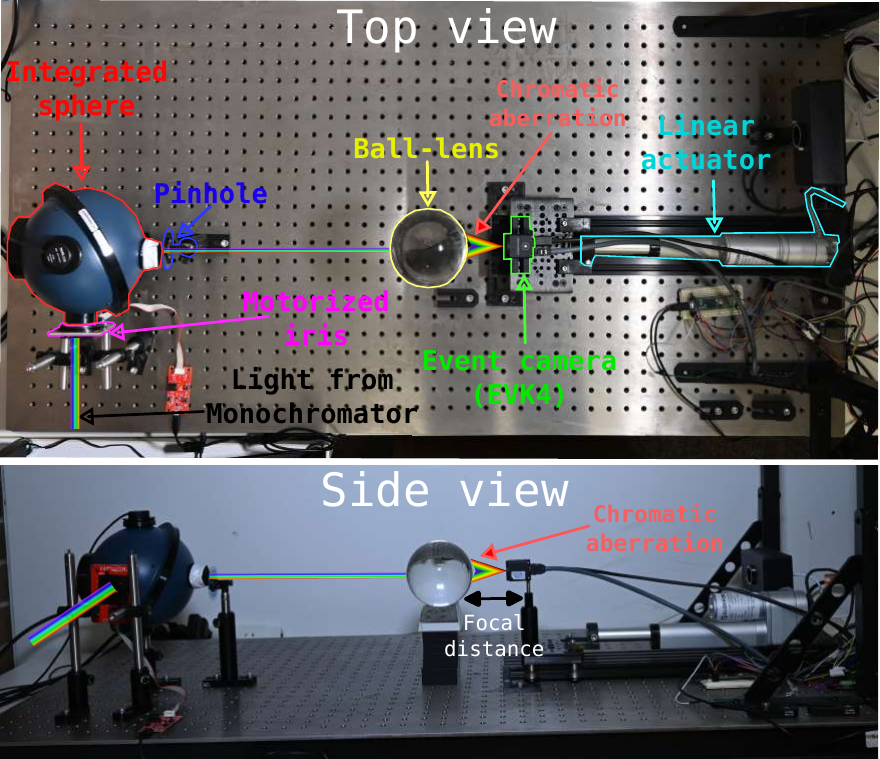}
  \caption{A top-side view of the optical setup used to characterise the spectral performance of the frame and event camera for visible light to near-infrared wavelengths.}
  \label{fig:optical_setup_real}
\end{figure}

To characterise the effect of chromatic aberration introduced by a ball lens for spectral analysis, we built the optical setup depicted in Figure~\ref{fig:optical_setup_real}. Data collection utilised two sensors: a Prophesee EVK4 (an event camera with logarithmic response to temporal contrast changes) and a monochrome Sony IMX249 sensor (a linear frame sensor)\footnote{https://en.ids-imaging.com/store/u3-3262se-rev-1-2.html}.

A 100 mm diameter K9 glass ball lens was illuminated by a narrow spectral band point source. We used a Newport TLS260B tunable light source\footnote{\url{https://www.newport.com/f/cs130b-configured-monochromators}}, generating monochromatic light between 400 nm and 1000 nm using a 280$\mu m$ slit output port. The monochromator output was coupled to an integrating sphere via a motorised Thorlabs iris, adjustable from 1 mm to 11 mm. The role of the iris is to ensure uniform photon count across the tested spectral range. The output of the sphere, is fixed with a pinhole with a 0.5 mm diameter and placed at 50 cm from the ball lens.

Each camera positioning relative to the ball lens was precisely controlled using a linear actuator driven by a Teensy microcontroller. For the EVK4 event camera data collection, a synchronised trigger signal was provided, enabling precise alignment of focal distances with event data. This synchronisation generated an additional data field, \( f_i \), representing focal distance measurements associated with each event, structured as \( e_i = \{x_i, y_i, p_i, t_i, f_i\} \). Such synchronisation was not needed for the grey level sensor for the frame sensor due to its discrete mode of operation.

We conducted the experiment in a dark environment, running the illumination source across the wavelength range of 400 nm to 1000 nm in increments of 50 nm. Power measurements for each wavelength were taken using a Thorlabs PM100 power meter positioned at one of the integrating sphere's outputs. All power measurements were set to values similar to the lowest expected photon count at 1000 nm wavelength, a band of lowest quantum efficiency (QE) and low monochromator emission, using the integration sphere input iris. Though the absolute power of the point source wasn't measured, all point sources can be treated as equal in every aspect but wavelength.

For each measurement, the selected wavelength was fixed, and the linear actuator oscillated the camera position from 0.5 mm to 30 mm relative to the ball lens at a frequency of 0.06 Hz. This frequency was chosen to accommodate the frame camera's required exposure time of 2 seconds, ensuring sufficient data capture. Moreover, this provided adequate temporal resolution for the event camera to record events at optimal focal positions for each wavelength. To ensure the reliability of the data and to account for mechanical and electronic inconsistencies in actuator movement, the linear actuator performed 50 forward-backwards cycles per wavelength, with only the central 30 cycles selected for event camera analysis. For the frame camera, extended exposure times and an additional 5-second waiting period between frames ensured stable, blur-free images, removing the need for repeated cycling. Analysis of the collected data was performed using the methodology detailed in Section 5.

\section{Ball Lens Simulation via Ray Casting}\label{sec:simulator}

\begin{figure}[h]
  \centering
  \includegraphics[width=\linewidth]{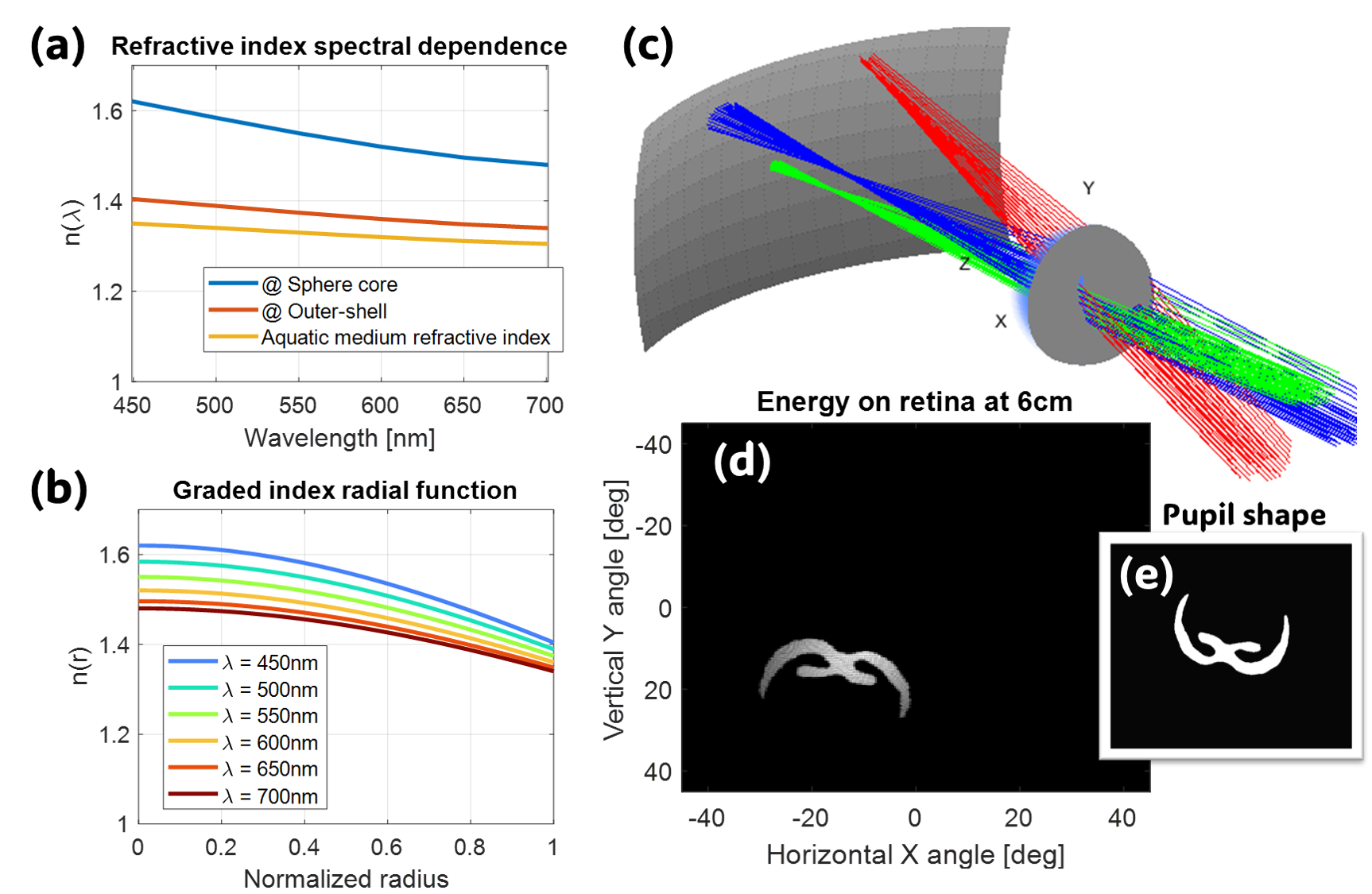}
  \caption{Ray casting simulation emulating cuttlefish vision. (a) Spectral dependence of core refractive index (blue line), outer sphere shell (red) and medium (yellow). (b) Graded index function of radius for various wavelengths. (c) Example ray tracing for three different point sources, through a pupil with the shape given in (e) and inspired by a Dwarf Cuttlefish pupil shape, then through a graded index sphere and finally reaching a retinal surface. The accumulated ray casting (50k rays) on an out-of-focus retina for a single point source.}
  \label{fig:simulation_example}
\end{figure}

To explain the ability for decomposition in a broad-spectrum environment, as most typically are, we explore other aspects of the cephalopod imaging system through the analysis of a ball lens transfer function coupled to various pupil shapes, including those unique to cephalopods. We do this by analysing changes in energy on the retina, according to point source ray casting onto a hemispherical retina for various pupils, wavelengths, and retina distance from the lens. 

A set of rays is emitted toward the ball lens sphere with with even angular spacing, where each ray encounters a binary pupil function, overlaid on the sphere surface, dictating its transmission. Inside the sphere, the refractive index follows a graded index (GRIN) profile:
\begin{equation}
{n(r,\lambda)=\frac{n_{\mathrm{core}}(\lambda)}{1+\bigl{(}n_{\mathrm{core}}(\lambda)/n_{\mathrm{outer}}(\lambda)-1\bigl{)}r^{2}/R^{2}}},\end{equation}
where $n_{core}$ is the refractive index in the centre of the sphere, and $n_{outer}$ is the refractive index on the outer surface of the sphere, both being a function of the wavelength, $\lambda$. In addition, we consider the refractive index of the aquatic medium, $n_{0}(\lambda)$. In this simulation, the specific combination of refractive index at every wavelength is the only manifestation of the wavelength choice, thus neglecting diffraction effects on the retina.

The direction of the refracted ray crossing the pupil, $e_{1}$, is calculated according to the vector form of Snell’s law:
\begin{equation}
{e_{1} = \tfrac{n_{0}}{n_{outer}}e_{0} + \bigl{(}\tfrac{n_{0}}{n_{outer}}cos\theta_{i} - cos\theta_{r}\bigl{)}e_{n1}},\end{equation}
where $e_{0}$ is the unit vector of the impinging ray, $e_{n1}$ is the unit vector of the normal to the sphere surface at the point where the ray intersects with the sphere, and the $\theta_{i}$ and $\theta_{r}$ are the impinging and refracted angles to that normal vector, respectively, and their relation is derived from Snell’s law:
\begin{equation}{sin\theta_{r} = \tfrac{n_{0}}{n_{outer}}sin\theta_{i}}.\end{equation}
To estimate each ray path inside the sphere, we use a numerical approximation according to the Euler method in Cartesian coordinates,

\begin{equation}{X_{i+1} = X_{i} + e_{i}ds},\end{equation}
\begin{equation}{e_{i+1}=e_{i}+\tfrac{1}{n_{i}}\nabla n ds},\end{equation}
where $ds$ is the simulation step, $X_{i}$ is the location of the ray in the sphere at each iteration, $e_{i}$ is the unit vector of the ray direction at each iteration (normalised to unit length for each simulation step), $\nabla n$ is the graded index gradient, and $n_{i}$ is the local refractive index value at location $X_{i}$.

At the outer radius of the sphere, rays exit into the external medium (assumed to also have $n_{0}(\lambda)$ refractive index values), following a similar process to the refraction of the entering ray with the reversed values of the refractive index values. The rays then propagate freely to the retinal surface, modelled as concentric hemispherical detectors positioned at different distances $R_{retina}$ from the sphere centre.

The energy distribution across these retina surfaces are determined by summing the contributions of all rays reaching an angular "pixel" location. Accumulating for various wavelengths, according to a chosen spectral composition of the source, forms an effective spatial-spectral mapping. As mentioned in previous sections, motion plays a crucial role in the perception process, and we may analyse various spatial-temporal-spectral mapping for various motion types, such as object motion or active scanning of the eye by changing $R_{retina}$ (sometimes called allocative perception\cite{zweifel_defining_2020} – where the internal sensor activity acts to expand its information gathering process). The simulation doesn't replicate the cephalopod eye exactly—where the ball lens moves axially against a static retina—our approach of flipping through concentric retina images captures a similar representation of the optical effects taking place in the sensor and may be used to explain cephalopod spectral perception.

To map the data representation to fit the spike-based animal retina, we input temporal contrast changes in retina intensity to an event-sensor simulator. This simulator uses pixel-level analysis to produce event streams from video input. While not encapsulating the full temporal resolution achievable by spike-based neural receptors, this method allows for emulating a spatial-temporal sparse representation of the captured scene and interpreting how the animal’s perception might deduce spatial-temporal-spectral information.

\section{Results and Analysis}\label{sec:Results}

Below, we characterise wavelength-dependent focal points identified through focal distance changes, using simulation (Section~\ref{sec:simulationanalysis}) and experimental optical setups (Section~\ref{sec:resultsspectral}), with both frame and event cameras. We further evaluate spectral discrimination performance (Section~\ref{sec:spectralperformane}) of the event camera across the visible spectrum.

\subsection{Spectral Characterisation via Simulation}\label{sec:simulationanalysis}

\begin{figure}[h]
  \centering
  \includegraphics[width=\linewidth]{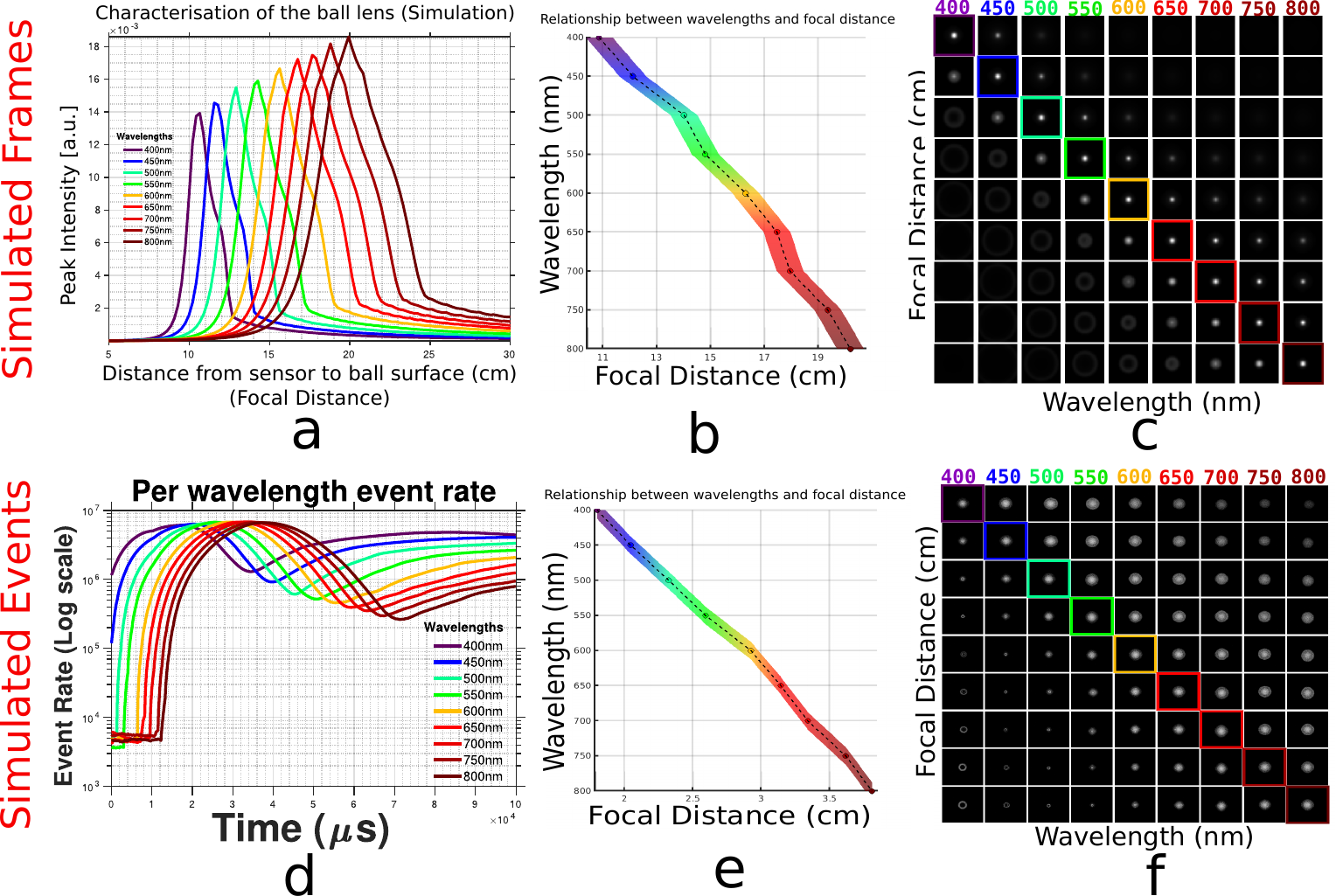}
  \caption{Spectral response results from the ray-casting simulator and the simulated events using~\cite{joubert_event_2021}. (a) Peak intensity profile as a function of focal distance for each wavelength. (b) and (e) Highlighting the relationship between wavelengths and focal distances from the frame and event camera, respectively. (c) and (f) PSF images across all focal distances and wavelengths, illustrate a clear diagonal trend where each wavelength comes into focus sequentially from the frame and event camera, respectively. (d) Event rate profile as a function of time for each wavelength, showing how each peak varies for different wavelengths.}
\label{tb:simulationanalysisresults}
\end{figure}

In this section, we present the final analysis of the data generated by the simulator detailed in Section~\ref{sec:simulator} for both frames and events. For each illumination point source, we send 500k rays (only some pass the pupil threshold) to a sphere of $3cm$ radius with refractive index properties as outlined in figure ~\ref{fig:simulation_example}(a) and (b). The retinal intensity maps are recorded for $R_{retina}=5cm$ to $R_{retina}=15cm$. Figure~\ref{tb:simulationanalysisresults}(a) illustrates the maximum intensity of each simulated PSF across the visible wavelength range (400 nm to 800 nm). The intensity peaks increase with wavelength, clearly showing an increase in the depth of focus at longer wavelengths. This occurs because the focal points shift further from the lens, causing more light rays to converge at these focal points.

The relationship between wavelength and focal length is nearly linear, exhibiting only minor variations in the focal distance with increasing wavelength (Figure~\ref{tb:simulationanalysisresults}(b) and (e) for frames and events, respectively). Furthermore, the width of the maximum pixel intensity broadens at longer wavelengths, indicating reduced spectral resolution and making it much harder to differentiate between adjacent wavelengths. Given the spectral distribution in seawater typically skews towards shorter (blue) wavelengths in the context of the cephalopod vision, higher spectral resolution at shorter wavelengths is advantageous.

This behaviour is further evident in the diagonal PSF outputs shown in Figure~\ref{tb:simulationanalysisresults}(d) and (f). While all wavelengths achieve focus, longer wavelengths exhibit significantly greater PSF contrast, which generates more events. It is clear that the PSF focuses and defocuses much faster toward the blue and green, and it slows down for the red wavelength, which indicates a shorter distance between the focal point at longer wavelengths and a gradual reduction in the spectral discrimination.

\subsection{Spectral Characterisation via the Real Setup}\label{sec:resultsspectral}

\begin{figure}[h]
  \centering
  \includegraphics[width=\linewidth]{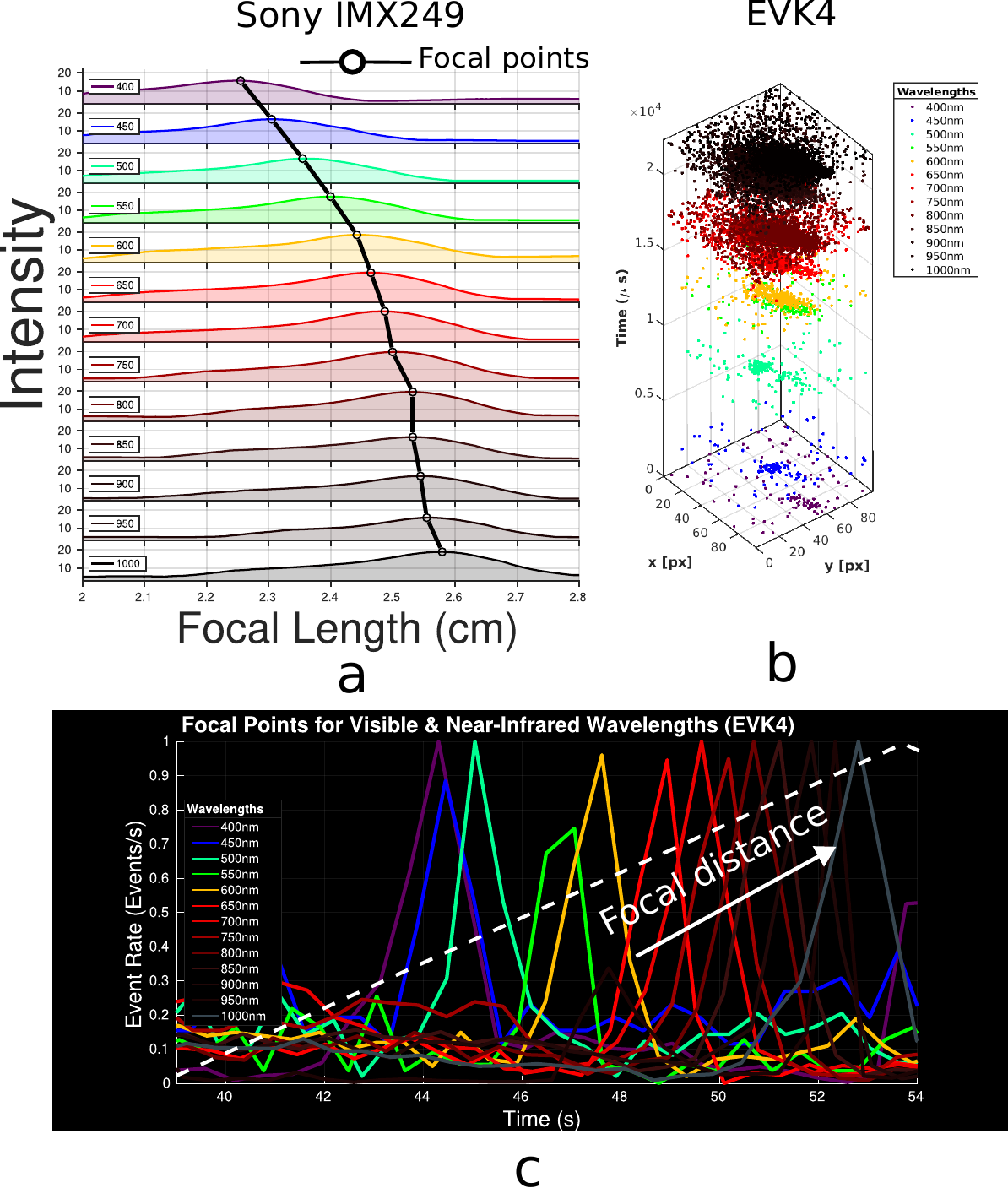}
  \caption{Spectral response measured using frame and event cameras in a real optical setup across visible and infrared wavelengths. (a) Peak intensity profile vs. focal distance for each wavelength, with the optimal focal positions indicated by the solid black line. (b) Colour-coded event point cloud illustrating focal positions from blue (400 nm) to near-infrared (1000 nm). (c) Peak event rate (normalised) as focal distance increases (moving away from the ball lens), showing sequential focusing from shortest wavelength (400 nm) to longest (1000 nm).}
\label{tb:realsetupresults}
\end{figure}

\begin{figure}[h]
  \centering
  \includegraphics[width=\linewidth]{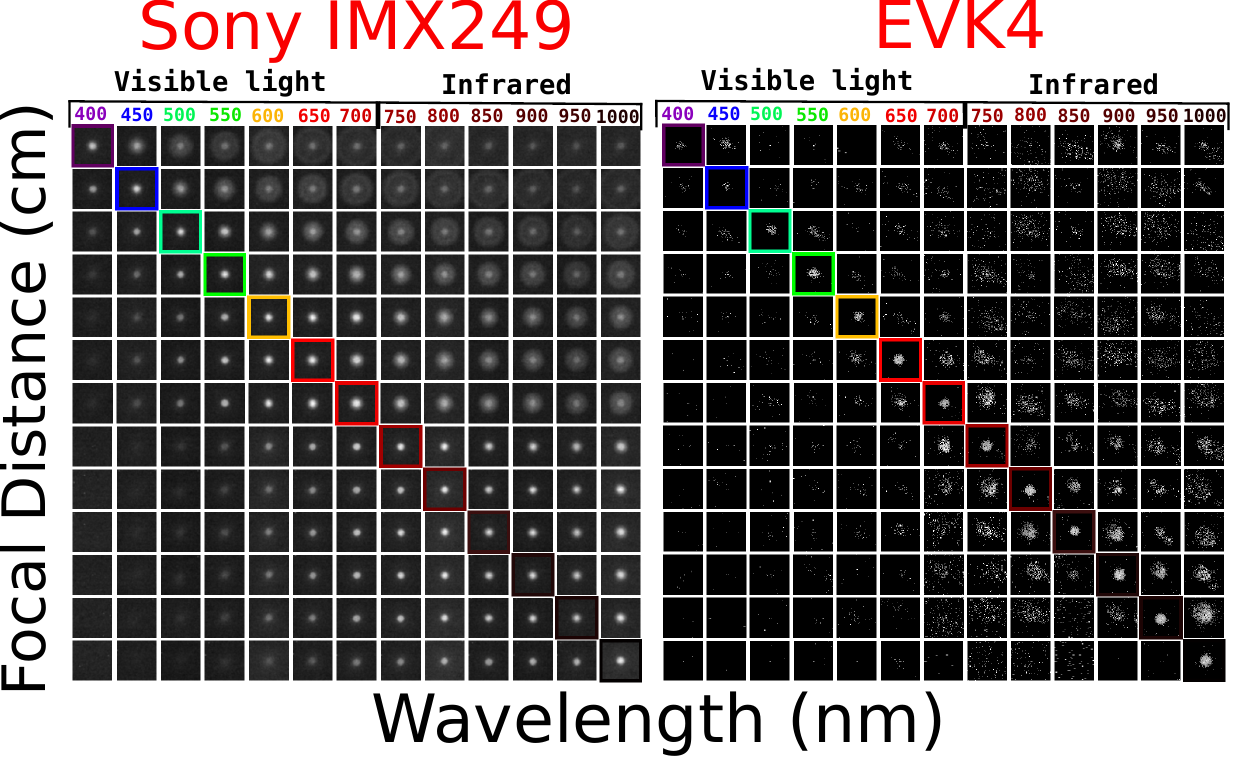}
  \caption{Comparison between the PSF focusing as a function of the wavelengths and focal distances from Sony IMX249 (Left) and the EVK4 (Right). The optimal focus is shown diagonally.}
  \label{fig:new_pinholes_visir_eventframes}
\end{figure}

As detailed in Section~\ref{sec:realsetup}, images were captured across various wavelengths and focal distances. For each image, the primary interest was the PSF intensity for the frame camera and the PSF event rate for the event camera.

Using the frame camera, we calculated the maximum pixel intensity within each image, which reveals distinct intensity peaks at different focal positions for each wavelength as shown in Figure~\ref{tb:realsetupresults}(a). This clearly illustrates chromatic aberration effects. By plotting wavelengths against the corresponding focal lengths, we observed a clear non-linear relationship that increased starting from the red wavelengths.

This non-linearity indicates that differences in focal distance become progressively smaller with longer wavelengths. In this case, distinguishing focal points at longer wavelengths, particularly from red to near-infrared, becomes challenging, negatively affecting spectral resolution at these longer wavelengths. To accurately capture these subtle differences and avoid skipping crucial wavelengths, more precise mechanical adjustments in focal positioning are necessary. This was also notably visible in Figure~\ref{fig:new_pinholes_visir_eventframes}(Sony IMX249), where the PSF quickly blurs after passing the focal point for shorter wavelengths but exhibits slower changes for near-infrared wavelengths.

Using the event camera, we observed an increase in event rates at wavelengths beyond 650 nm, extending into the near-infrared range, Figure~\ref{tb:realsetupresults}(b). This rise in events contributes to increased noise within the event stream, reducing spectral resolution. This shows that near-infrared wavelengths generate more events than blue wavelengths, even with a calibrated light source. This wavelength-dependent increase in event rate was also observed by simulations (Figures~\ref{tb:simulationanalysisresults}(a) and (b)). Additionally, as shown in Figure~\ref{tb:realsetupresults}(c), the spacing between the event-rate peaks decreased non-linearly at longer wavelengths.

As illustrated in Figure~\ref{fig:new_pinholes_visir_eventframes}(EVK4), the diagonal events within the focused PSF appear faintest at shorter (blue) wavelengths, but the event rate increases at longer wavelengths, despite maintaining calibrated wavelengths. This increase in intensity/event rate at higher wavelengths results from the increase in depth-of-focus, which effectively concentrates more photons at the focal point for longer wavelengths. These findings are also closely aligned with the simulated results in Section~\ref{sec:simulationanalysis}.

In the context of the cephalopods, near-infrared wavelengths are typically irrelevant because they do not significantly penetrate seawater, and the underwater spectral distribution is predominantly skewed towards shorter (blue) wavelengths.

\subsection{Performance on Colour Discrimination}\label{sec:spectralperformane}

We assess the performance of colour discrimination by measuring the spectral discrimination ratio. We proposed a dimensionless spectral resolution metric $\eta$ to measure the system's ability to discriminate between wavelengths. A higher value of $\eta$ indicates higher spectral discrimination, meaning the system can resolve finer spectral details.

\begin{figure}[h]
  \centering
  \includegraphics[width=\linewidth]{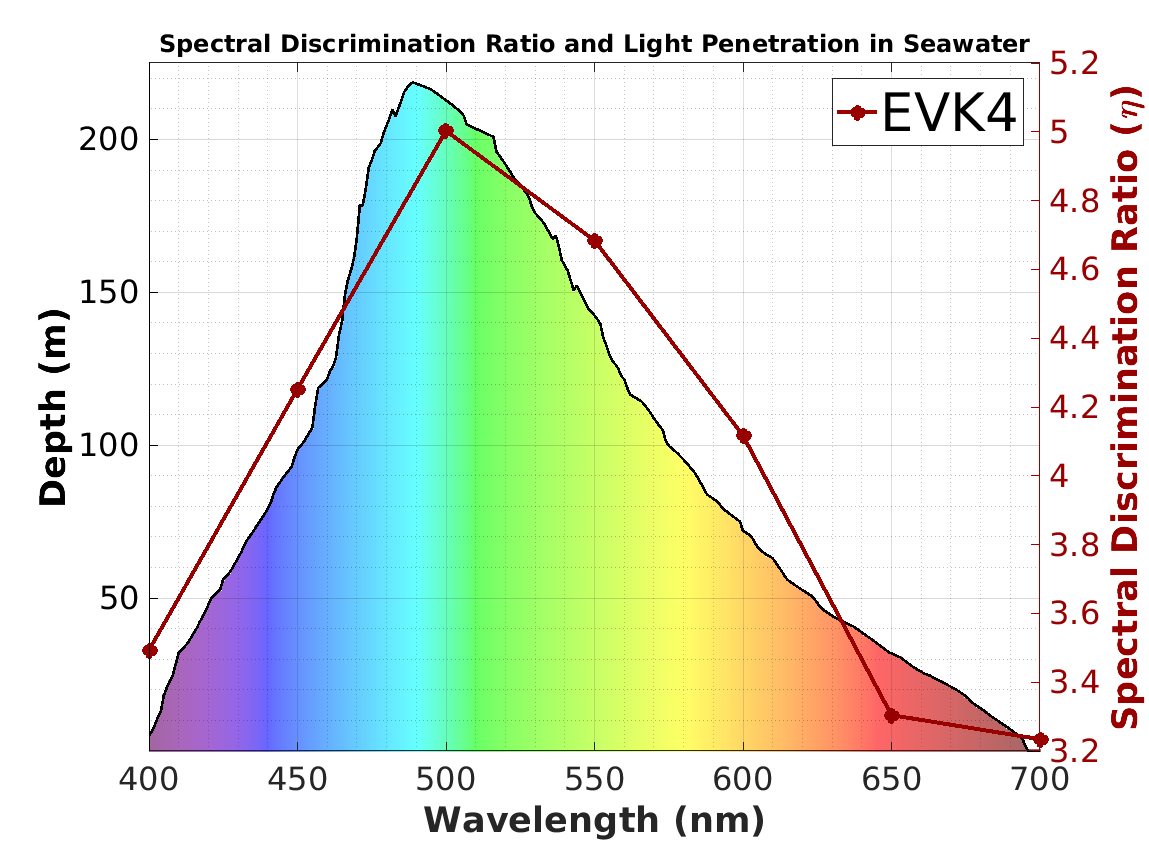} %
  \caption{Spectral discrimination ratio. The coloured under-curve shows the light penetration profile over the visible spectrum (400–700 nm). The right y-axis displays the spectral discrimination ratio, $\eta$, from the EVK4 event camera, revealing higher resolution in the blue-green region and lower resolution toward red wavelengths.}
  \label{fig:spectral_resolution}
\end{figure}

As described in Section~\ref{sec:simulationanalysis} and Section~\ref{sec:resultsspectral}, for each wavelength the system produces an intensity profile as a function 
of the focal distance, \( f \). The peak of the profile is located at \( f_0 \), 
and the full width at half maximum (FWHM) of the profile, FWHM, 
represents the spread of the response.

Since the mapping between focal distance and wavelength, \(\lambda\), is considered nonlinear (Figure~\ref{tb:realsetupresults}(b) and Figure~\ref{tb:realsetupresults}(e)), we convert FWHM into a spectral width, \(\Delta \lambda\), using the local derivative:

\begin{equation}
\Delta \lambda = \left.\frac{d\lambda}{df}\right|_{f_0} \cdot FWHM.
\end{equation}

The spectral resolving power (i.e spectral discrimination ratio), $\eta$, is then defined as:
\begin{equation}
\eta = \frac{\lambda}{\Delta \lambda}.
\end{equation}

$\eta$ (unitless) is obtained from the experimental results using the EVK4 sensor and is presented in Figure~\ref{fig:spectral_resolution}. The results highlight an interesting and consistent pattern: simulation and the real sensor demonstrate significantly higher spectral discrimination within the blue-green wavelength range, while discrimination capabilities notably decrease towards the red wavelengths. This is also consistent with~\cite{chung_complex_2017} for coastal cephalopods that have a high spectral sensitivity at 484–505 nm.

This was also observed in Figure~\ref{tb:simulationanalysisresults}(a) and Figure~\ref{tb:realsetupresults}(a), where the intensity profile's width and height expand with increasing wavelength, which indicates a lower spectral discrimination at higher wavelengths. We limited the analysis to 700nm, consistent with cephalopod colour perception~\cite{jagger_wide-angle_1999} limit. This is also justified by the significant absorption of infrared wavelengths in seawater, making them irrelevant for cephalopods.

The pronounced spectral sensitivity within the blue-green spectrum range is particularly beneficial for coastal cephalopods due to the increased dominance and penetration of these wavelengths. At a much deeper level, deep-sea species often rely on bioluminescence as the only light source~\cite{otjacques_bioluminescence_2023}. These findings show how the experimental setup mirrors the spectral sensitivity of coastal cephalopods.

\section{Discussion and Future Work}

The ability to distinguish between sparse narrow-bandwidth signals, using motion and chromatic aberration, is evident from the quantitative results of both the experiment and simulation. It may be argued that the aquatic environment is adequately sparse in a certain spatial representation. We expect limited sharp contrasts in space (object edges) and slowly varying spectral distributions, like in terrestrial environments, and becoming more so for creatures in a deep-sea habitat. However, a cuttlefish living in the shallows, with ample sunlight and colourful (broad spectrum) environments, will find colour separation using only concentric radially uniform PSFs not as effective.

Here comes the unique pupil shapes into play, with off-axis light entering only from selective regions. Distinct patterns on the retina caused by these narrow pupils have been suggested to allow for colour separation~\cite{stubbs_spectral_2016}. When expanding the concept to motion-based analysis, we can see a direct translation from PSF evolution to impinging spectra from a point broadband source.

\begin{figure}[h]
  \centering
  \includegraphics[width=3.1in]{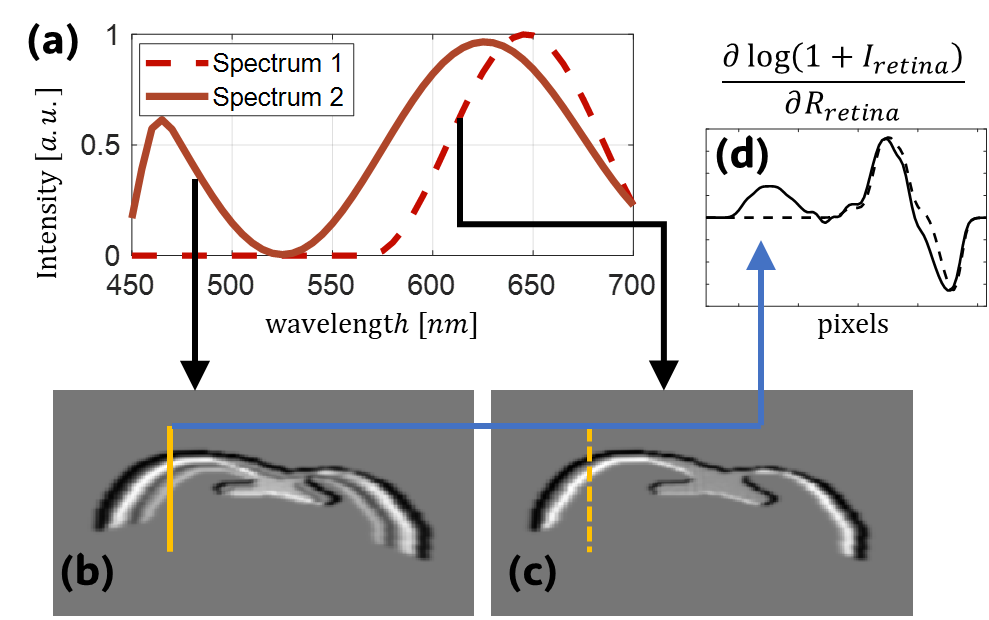}
  \caption{Spatial-spectral reconstruction. (a) Two broad-spectrum examples are used for point sources in the simulation framework, where spectrum 1 contains only a long wavelength band (dashed line) and spectrum 2 is both short and long wavelength (solid line). The gradient along the retina-lens axis of the retina log intensity map is computed and presented for spectrum 1 point source (b) and spectrum 2 point source (c).
The profiles slicing the side streaks of each such gradient image (d) can be directly related to the original spectrum of the point source.}
  \label{fig:discussion_fig}
\end{figure}

We look at an example of two impinging spectra, Figure~\ref{fig:discussion_fig}(a). Both seem red to a human observer, even though spectrum 2 contains some short-wavelength bands. The overall retinal illumination was calculated for these sources for a narrow pupil shape, and we plot here the gradient along the radial axis of the log-intensity map on the retina caused by a purely red spectrum 1 Figure~\ref{fig:discussion_fig}(b), and for an identical point source, but with a mixture of red and blue Figure~\ref{fig:discussion_fig}(c). By plotting the gradient along this axis, we are evaluating the log-intensity change with the axial motion of the retina-lens distance, which can be related to spike generation rates across the retina (or an event sensor). Examining the cross-section of the streaks or side-lobes of the shape in this representation, we see a pattern that can be directly associated with the impinging spectra for each source, Figure~\ref{fig:discussion_fig}(d). Evaluating a more accurate spectral decomposition metric, together with a computational method to extract spectral images from extended targets (and non-sparse scenes), is a work in progress.

In future work, we will also explore techniques for rapidly varying the focal length automatically using devices such as variable-focus liquid lenses for more continuous spectral information extraction. Having demonstrated system performance under controlled laboratory conditions with discrete wavelengths, our next step is to evaluate spectral discrimination capabilities in natural environments on objects with multiple wavelengths.

\section{Conclusion}
In this paper, we presented a cephalopod-inspired approach that leverages chromatic aberration from a ball lens to capture spectral information across the visible and near-infrared ranges. By mimicking cephalopods' dynamic focal shifts through a motorised system, we demonstrated how event-based and frame-based cameras can directly capture spectral data, eliminating the need for colour filters or complex demosaicing, even for sparse, narrowband signals. Our ray simulator, which accounts for the ball lens's graded-index properties and the curved retina structure, was used to characterise the role of chromatic aberration and confirmed its ability to enable spectral discrimination. Using a motorised optical platform, we characterised the system's ability for spectral discrimination and were able to infer spectral information directly from event stream data. Notably, the system's spectral resolution fits the requirements imposed by cephalopods' habitat. The aquatic attenuation spectrum supports the notion of higher spectral resolution at shorter wavelengths, where colour cues may convey information regarding the distance and composition of the scene, while higher photon sensitivity in long wavelengths can help penetrate the lower-intensity signals at these bands. This work signals new directions for visual sensing systems driven by nature's evolutionary solutions and a novel approach to enhancing neuromorphic vision systems.

\textbf{Acknowledgement}. We thank Paul Hurley, Paul Kirkland and Jonathon Wolfe for providing insights and comments that greatly improved this work. We also thank Nadil Lekamge for 3D modeling and 3D printing of the linear actuator platform.

{\small
\bibliographystyle{ieee_fullname}
\bibliography{egbib}
}

\end{document}


\maketitle

In this document, we describe our synchronisation, light source calibration approach used in the paper "Seeing like a Cephalopod: Colour Vision with a Monochrome Event Camera" and show a simple technique to perform colour segmentation by focusing.

\section{Synchronisation and Data Acquisition Hardware}

\begin{figure}[h]
  \centering
  \includegraphics[width=\linewidth]{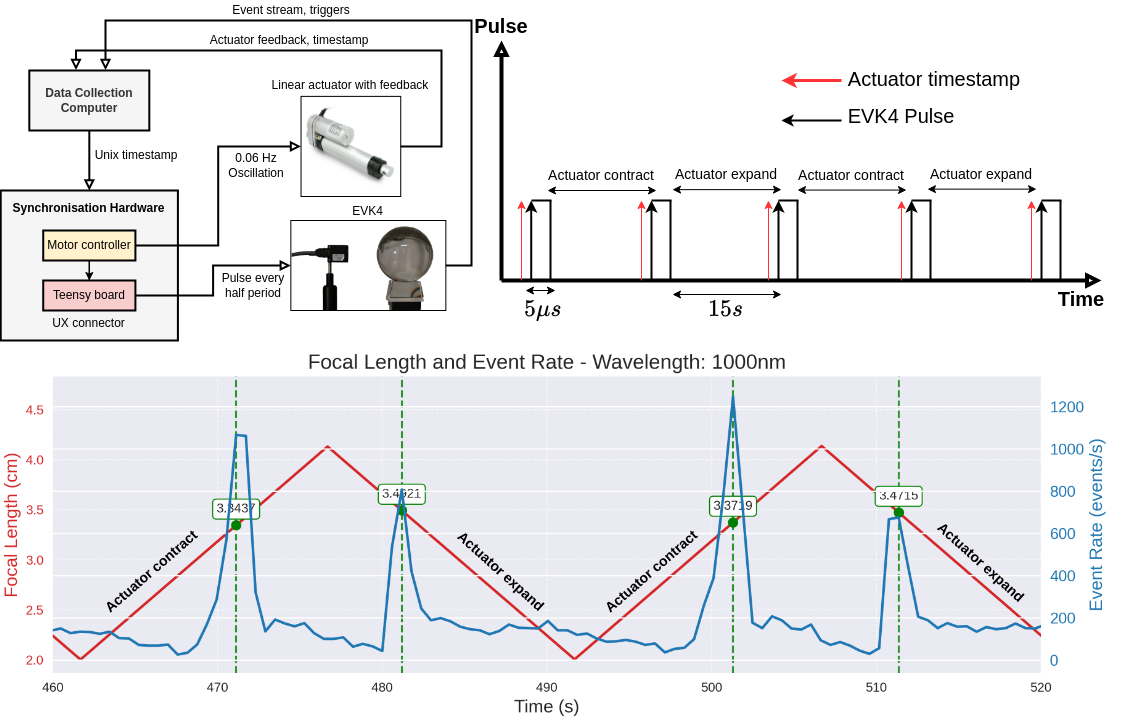} %
  \caption{The system architecture for data acquisition. \textbf{Top:} The event camera is synchronised with the linear actuator to obtain accurate focal distance values. \textbf{Bottom:} The measured event rate (blue) and focal distance (red) curves demonstrate the consistent peak in event rate over the same focal point.}
  \label{fig:synchronation_figure}
\end{figure}

Due to the event camera's high temporal resolution, accurate synchronisation is essential. In this study, synchronisation was achieved using a physical synchronisation signal between the host computer, linear actuator, and event camera (Figure~\ref{fig:synchronation_figure}).

The event camera's synchronisation cable (IX connector) connects to a Teensy board, which generates trigger pulses (rising and falling edges) each time the linear actuator completes a half oscillation ($\frac{f}{2}$). Since the Teensy lacks a Real-Time Clock (RTC), the host computer provides a Unix timestamp to the Teensy at the start of actuator movement. Both the timestamp and actuator feedback (in mm) are recorded.

The actuator was programmed to oscillate between 1 cm and 3 cm at 0.06 Hz. At each maximum extension and contraction, a synchronization pulse is sent to the event camera, and the corresponding Unix timestamp and actuator position are logged. Event data acquisition was performed using Gen4 software\footnote{https://github.com/neuromorphicsystems/gen4}, which automatically records trigger timestamps.

Due to initial misalignment between host computer and event camera timestamps, post-processing alignment was performed. We determined the consistent offset between trigger events and actuator timestamps and corrected all timestamps accordingly. The synchronization delay was approximately 10–50 ms, acceptable given the slow actuator frequency. The low frequency was chosen specifically because the event camera's response slows in dark environments, necessitating slower movements to generate sufficient event data.

Figure~\ref{fig:synchronation_figure} (bottom panel) demonstrates event rate peaks during actuator contraction and expansion, clearly indicating that at these focal positions, the camera focuses near-infrared (1000 nm) wavelengths while other wavelengths remain defocused.

\section{Light Source Calibration}

Because the camera spectral sensitivity depends on the quantum efficiency, the wavelength intensity from the monochromator~\footnote{\url{https://www.newport.com/f/cs130b-configured-monochromators}} needs to be adjusted to produce the same number of photons for the imaging sensor. The raw intensity (i.e. frame) is captured by the camera at a given wavelength $\lambda$ that is proportional to the product of the incident 
power $P(\lambda)$ and the QE, $\mathrm{QE}(\lambda)$. Because $\mathrm{QE}(\lambda)$ of both the Sony IMX249~\footnote{\url{https://en.ids-imaging.com/store/u3-3262se-rev-1-2.html}} and the EVK4 decay rapidly after 700 nm toward 1000 nm, directly measuring the same optical power at all wavelengths would result in a camera signal biased by the camera's spectral response (i.e. Quantum efficiency). To address this, we designate 1000\,nm (where the camera is least sensitive) as our reference. That is, we measure the 
"full-iris" power at 1000\,nm, $P_{\text{full}}(1000)$, and multiply by 
$\mathrm{QE}(1000)$ to determine the baseline camera signal for the dimmest 
region of the spectrum. By using this low-sensitivity endpoint as the reference, 
we avoid overexposure or saturation at shorter wavelengths. We then replicate 
this same "reference signal" at each other wavelength $\lambda$ by adjusting 
the incident power such that
\[
P_{\text{target}}(\lambda)\,\times\,\mathrm{QE}(\lambda)
\;=\;
P_{\text{full}}(1000)\,\times\,\mathrm{QE}(1000).
\]
In practice, we also measure and subtract a small "dark power" offset, 
$P_{\text{dark}}$, to account for baseline noise in our system. 
Specifically, the net photoelectron-generating signal becomes
\[
\Bigl[P_{\text{full}}(\lambda)\,\times\,\mathrm{QE}(\lambda)\Bigr]
\;-\;
P_{\text{dark}},
\]
and we adjust $P_{\text{full}}(\lambda)$ accordingly to match that obtained 
at 1000\,nm. By calibrating in this manner, the camera captures the same 
effective brightness at every wavelength, thereby ensuring meaningful, 
comparable data across the visible and near-infrared range. Further details, 
including the measured, adjusted, and final target power values, are provided 
in Table~\ref{tab:calibration}.

\begin{table}[ht]
\centering
\caption{Light source calibration values across visible and near-infrared.}
\begin{tabular}{@{}cccccc@{}}
\toprule
$\lambda$ (nm) & $\mathrm{QE}$ & $P_{\text{full}}$ (nW) & $P_{\text{dark}}$ (nW) & $P_{\text{adjusted}}$ = ($P_{\text{full}}$/$\mathrm{QE}$)-$P_{\text{dark}}$ (nW) & $P_{\text{target}}$ (nW) \\ \midrule
400   & 0.5805 & 36.28  & 0.02554 & 21.04  & 5.43  \\
450   & 0.7422 & 84.28  & 0.02554 & 62.53  & 4.24  \\
500   & 0.7852 & 147.4  & 0.02554 & 115.72 & 4.01  \\
550   & 0.7060 & 184.2  & 0.02554 & 130.03 & 4.46  \\
600   & 0.6088 & 198.2  & 0.02554 & 120.65 & 5.17  \\
650   & 0.5008 & 195.2  & 0.02554 & 97.74  & 6.29  \\
700   & 0.3892 & 164.2  & 0.02554 & 63.89  & 8.09  \\
750   & 0.2884 & 133.1  & 0.02554 & 38.37  & 10.92 \\
800   & 0.1966 & 116.9  & 0.02554 & 22.97  & 16.02 \\
850   & 0.1318 & 109.1  & 0.02554 & 14.36  & 23.89 \\
900   & 0.0795 & 135.6  & 0.02554 & 10.77  & 39.58 \\
950   & 0.0399 & 194.1  & 0.02554 & 7.72   & 78.93 \\
1000  & 0.0146 & 217.1  & 0.02554 & 3.15   & 215.39 \\ \bottomrule
\end{tabular}
\label{tab:calibration}
\end{table}

\section{Longitudinal Chromatic Aberration of the Ball Lens}

\begin{figure}[h]
  \centering
  \includegraphics[width=0.8\linewidth]{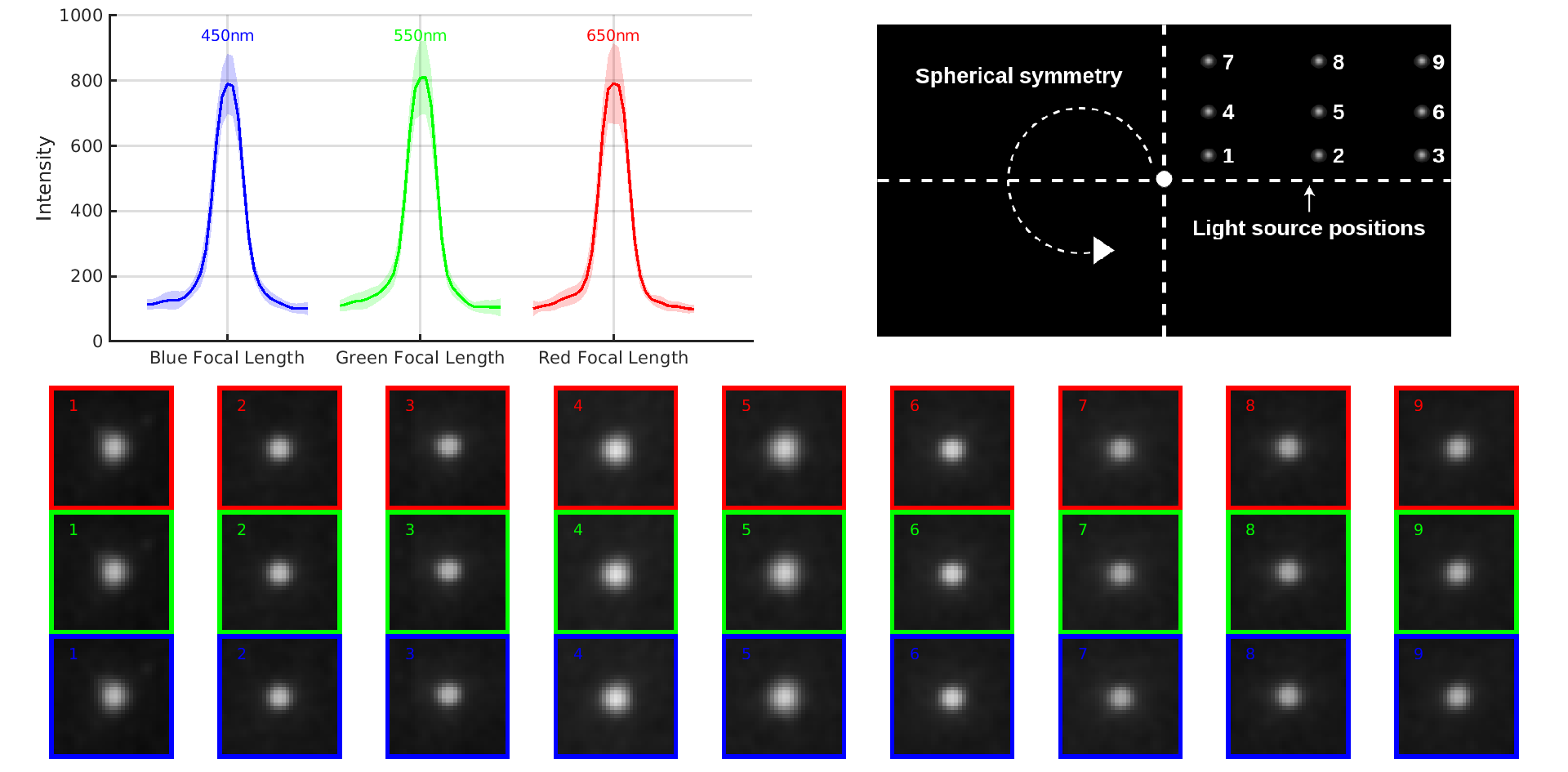} %
  \caption{Longitudinal chromatic aberration from the ball lens. Different points across the field of view produce similar chromatic aberrations effects, enabling consistent spectral discrimination at any position within the image.}
  \label{fig:longitudinal_chromatic_aberration}
\end{figure}

We investigated the type of chromatic aberration produced by the ball lens, focusing specifically on distinguishing between Longitudinal Chromatic Aberration (LCA, or Axial Chromatic Aberration) and Transverse Chromatic Aberration (Lateral Chromatic Aberration). Our experiment confirmed that the ball lens primarily produces Longitudinal Chromatic Aberration, where different wavelengths focus at distinct points along the optical axis, resulting in uniform aberration across the entire field of view regardless of the point source's position.

To validate this, we moved a pinhole to nine distinct positions within the first quadrant of the imaging field of view. Due to the spherical symmetry of the ball lens, the aberration pattern observed in one quadrant is identical in the other quadrants. Figure~\ref{fig:longitudinal_chromatic_aberration} illustrates the presence of LCA. For each position, we separately recorded images at three wavelengths (450 nm, 550 nm, and 650 nm) while adjusting the focal plane accordingly for each wavelength. A total of nine images per position (three wavelengths, each with three focal planes) were captured. The aggregated cross-sectional profiles for each wavelength, shown in Figure~\ref{fig:longitudinal_chromatic_aberration} (top left), highlight subtle yet consistent differences between wavelengths and demonstrate the lens's uniform aberration characteristics.

\section{Spectral (Colour) Segmentation Algorithm}

\begin{figure}[h]
  \centering
  \includegraphics[width=0.8\linewidth]{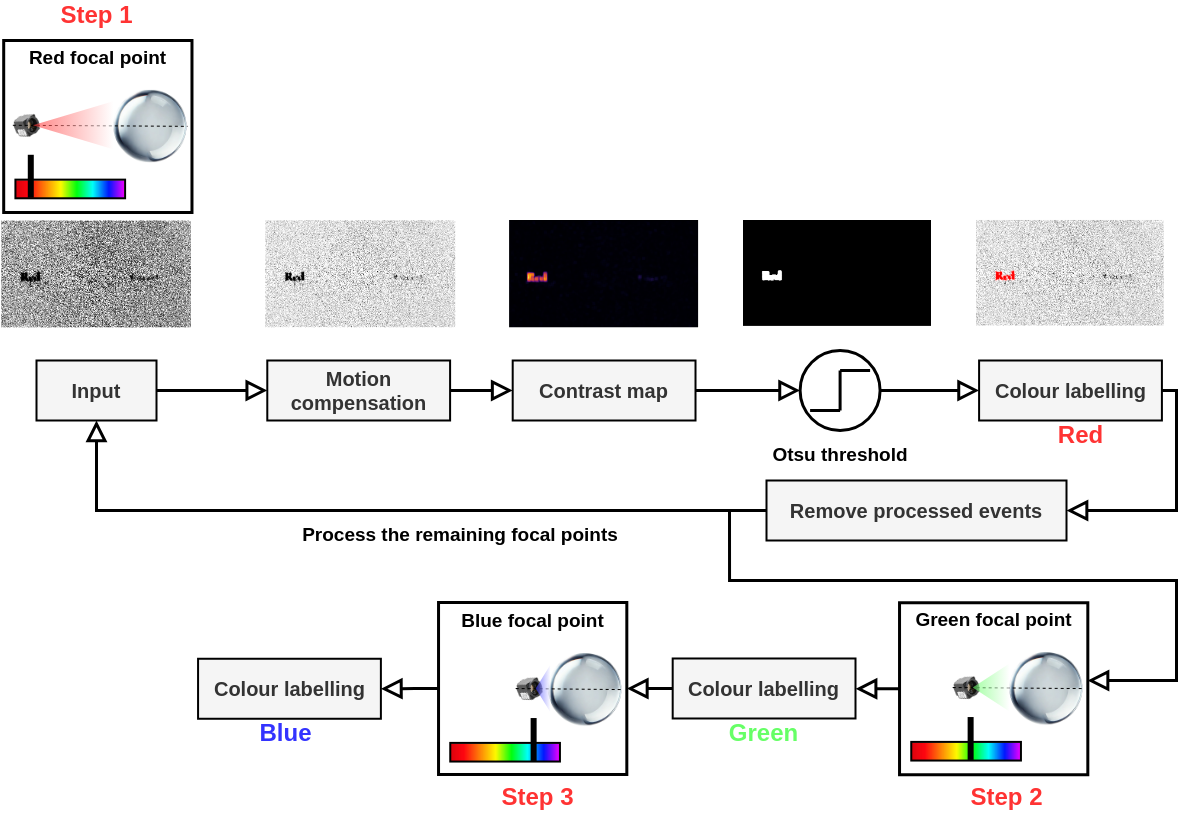}
  \caption{Colour segmentation algorithm: A step-by-step procedure to automatically label event streams based on local sharpness in the accumulated event image, effectively mapping events to their corresponding spectral information.}
  \label{fig:focal_segmentation_algorithm}
\end{figure}

We propose a simple algorithm (Figure~\ref{fig:focal_segmentation_algorithm}) to segment spectral information from event camera data. Our dataset was captured in an uncontrolled environment, where a cellphone screen displayed “Red,” “Green,” and “Blue” in their respective colours. The algorithm assumes known focal distances for specific wavelengths and shifts the sensor’s focal plane accordingly. When focused on a particular colour, events corresponding to that wavelength appear at distinct spatial locations.

Before running the segmentation, we perform a calibration step. Using a spectrometer, we identify the linear actuator positions that yield peak intensity (sharpest focus) for each wavelength. This process is repeated for all target wavelengths, resulting in an accurate actuator-to-wavelength mapping.

Once calibrated, the algorithm systematically moves the sensor back and forth through the defined focal ranges. During each sweep, it accumulates all triggered events into a single event image by counting events at each pixel. To enhance clarity, we then apply a Contrast Maximisation (CMax) algorithm~\cite{gallego_unifying_2018}, assuming linear motion, to effectively deblur and compensate for small movement.

Next, a 30×30 sliding window traverses over the accumulated image to measure local sharpness by calculating variance~\cite{gallego_focus_2019} within each window; higher variance indicates sharper focus. We binarise this variance map using Otsu’s thresholding to isolate the sharpest regions. Events in these regions are labeled according to their spectral content, then removed from subsequent processing. The actuator moves to the next focal plane, and the process repeats until all wavelengths have been segmented. The number of colour labels depends on how many wavelengths to focal points are known during the calibration step. In future work, we aim to develop a technique that doesn't require a calibration step to perform spectral segmentation in a single shot.

\begin{figure}[h]
  \centering
  \includegraphics[width=0.8\linewidth]{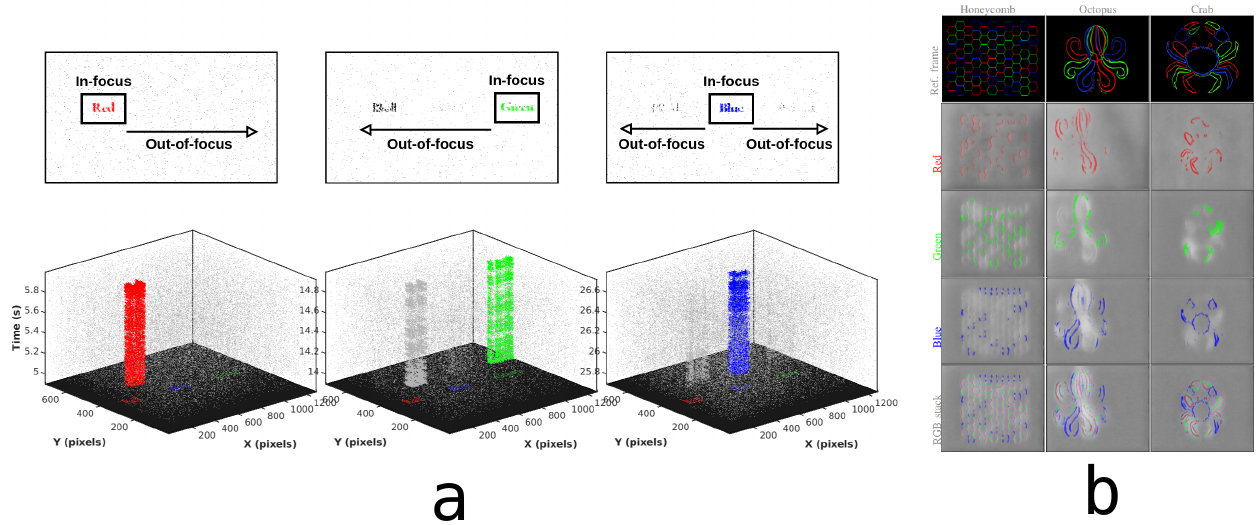} %
  \caption{Colour segmentation on 2D objects. (a) Segmentation of three objects that are separated from each other. (b) Segmentation of objects that are close to each other. Both show how the cephalopod-inspired event camera can perceive colours and how they are segmented.}
  \label{fig:Coloursegmentation2d}
\end{figure}

\begin{figure*}[h] 
\centering
\renewcommand*{\arraystretch}{0.3}
\setlength{\tabcolsep}{0.5pt}

\setlength{\fboxrule}{0.5pt}%
\setlength{\fboxsep}{0pt}   %

\begin{tabular}{c c c c c c c}
    \color{gray!90}Reference frame & \color{gray!90}T=1.5s (\color{red!90}Red) & \color{gray!90}T=10.2s (\color{red!90}Red) & \color{gray!90}T=17.8s (\color{green!90}Green) & \color{gray!90}T=21.8s (\color{green!90}Green) & \color{gray!90}T=27.6s (\color{blue!90}Blue) & \color{gray!90}T=34.6s (\color{blue!90}Blue) \\
    
    \fcolorbox{black}{white}{\includegraphics[height=0.76in,width=0.95in]{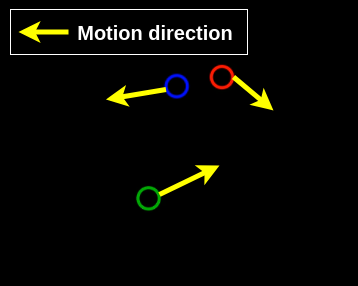}} &
    \fcolorbox{black}{white}{\includegraphics[height=0.76in,width=0.95in]{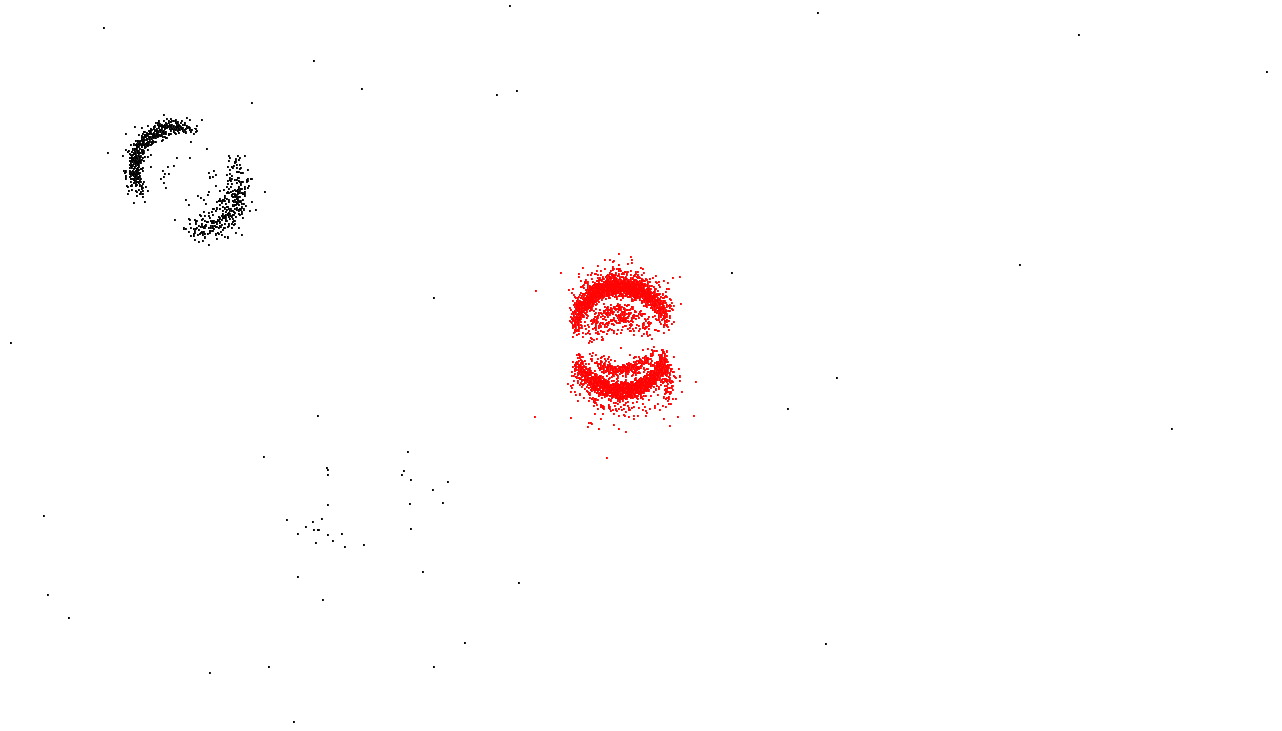}} &
    \fcolorbox{black}{white}{\includegraphics[height=0.76in,width=0.95in]{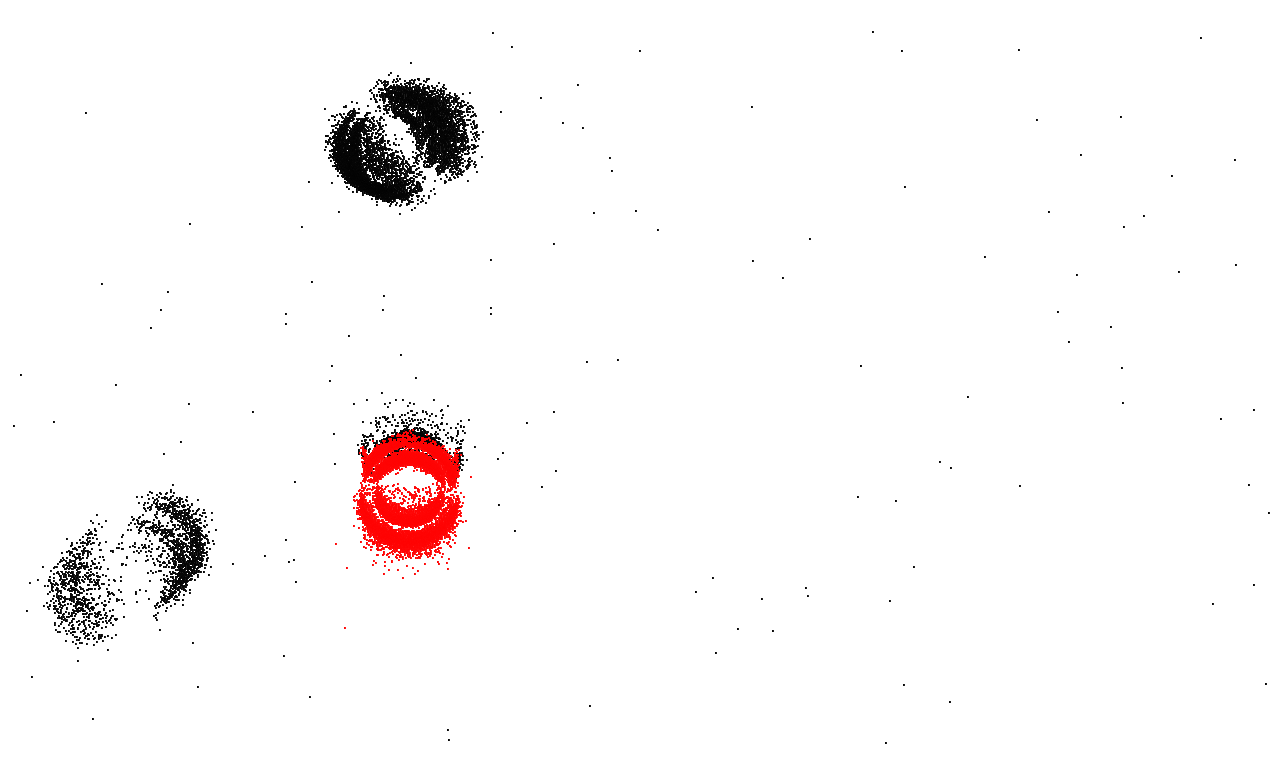}} &
    \fcolorbox{black}{white}{\includegraphics[height=0.76in,width=0.95in]{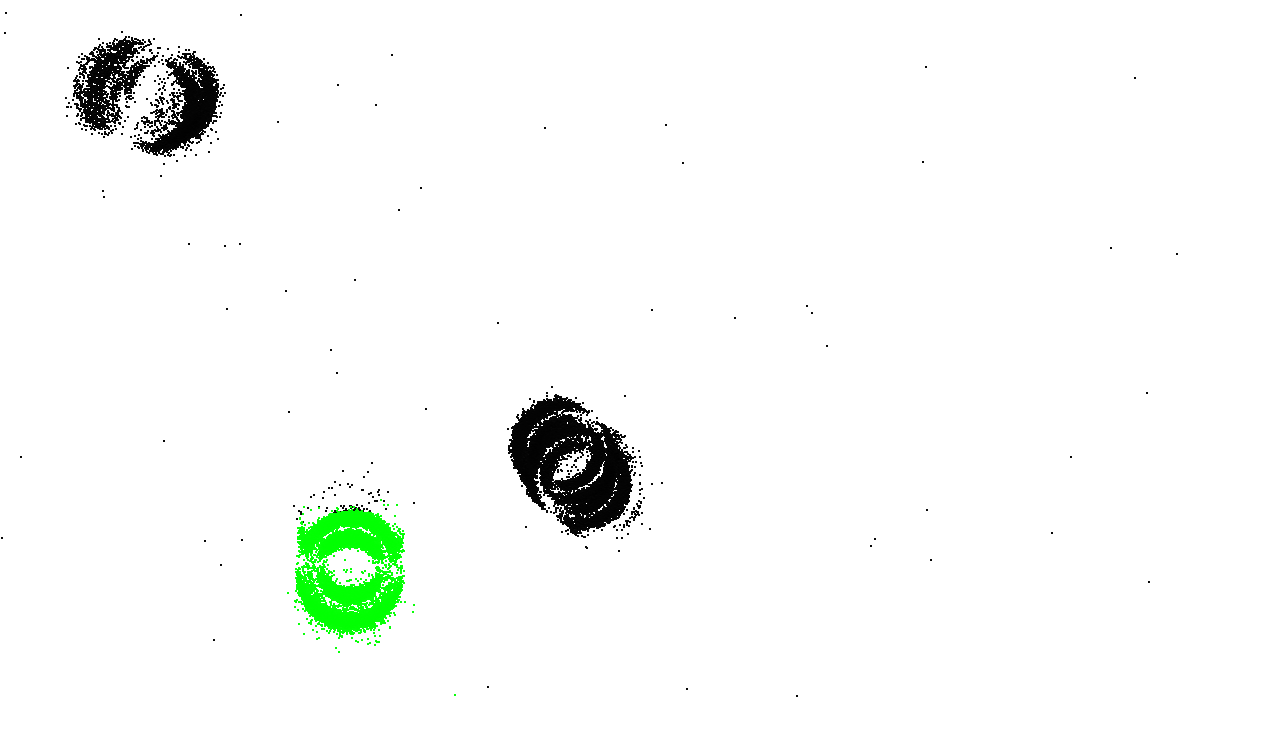}} &
    \fcolorbox{black}{white}{\includegraphics[height=0.76in,width=0.95in]{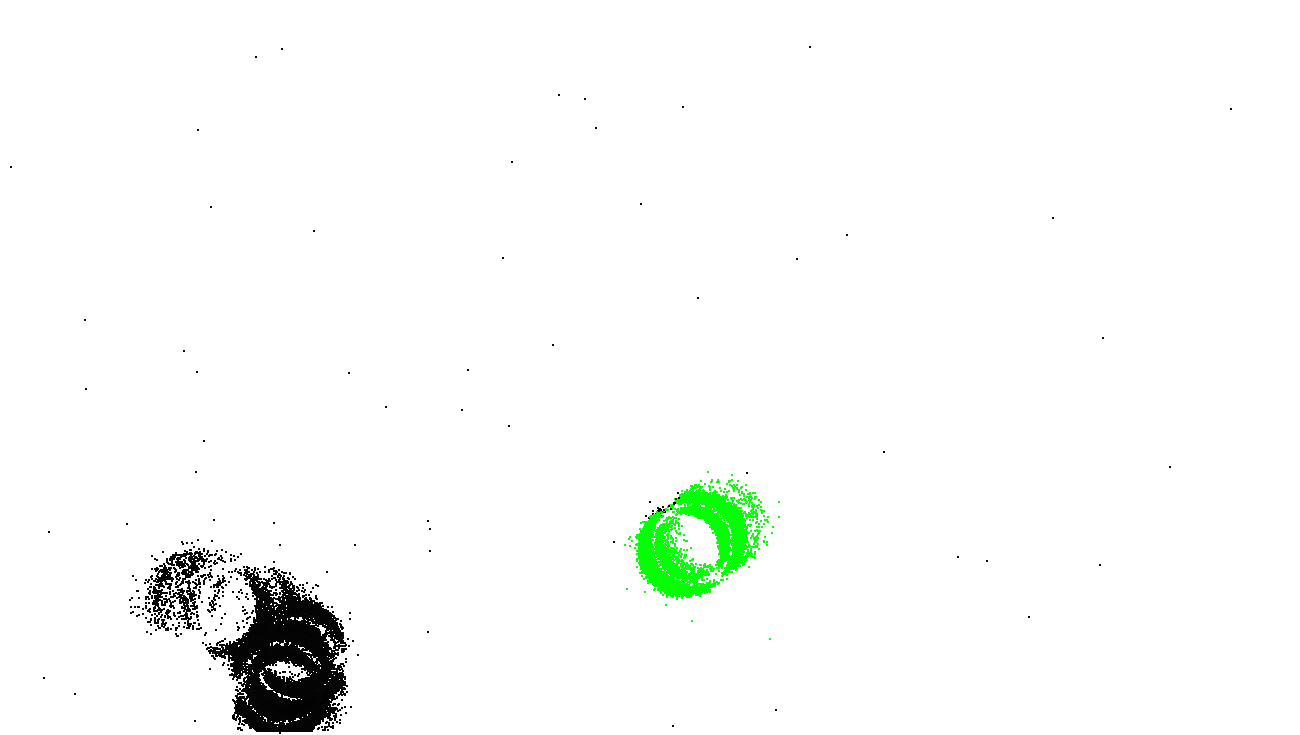}} &
    \fcolorbox{black}{white}{\includegraphics[height=0.76in,width=0.95in]{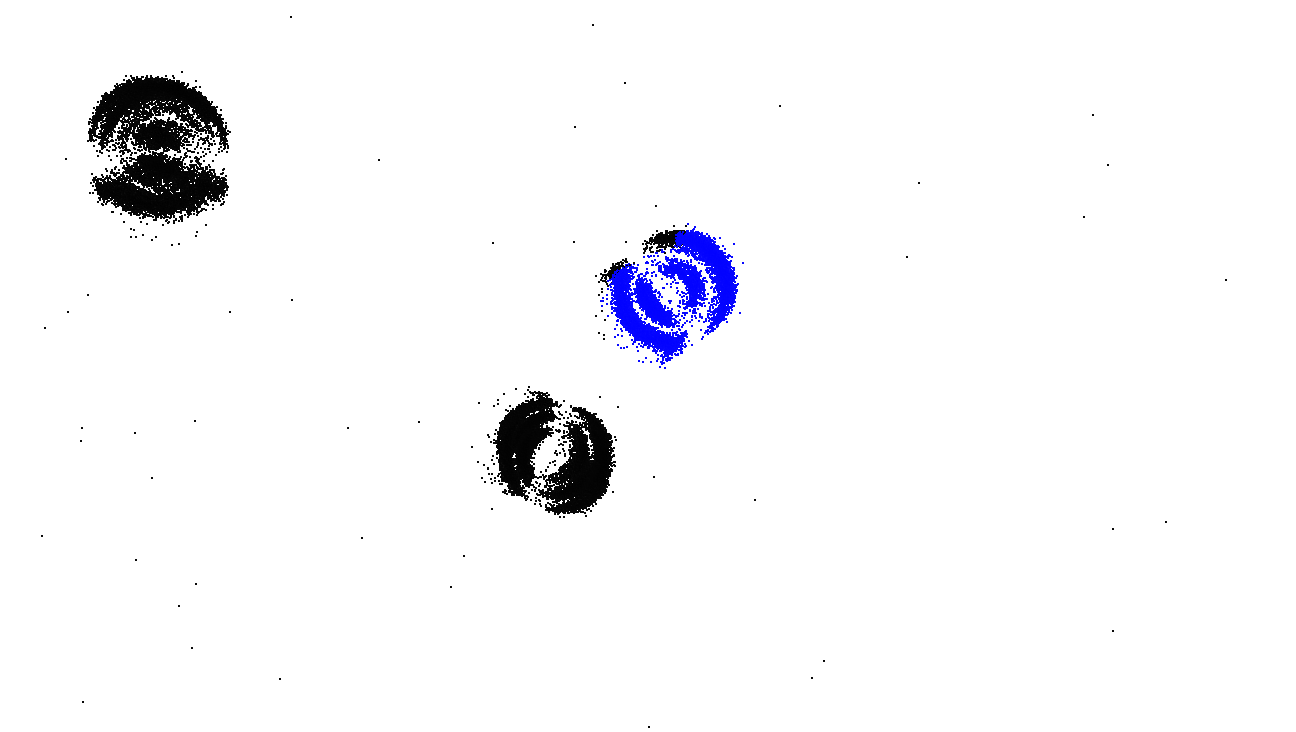}} &
    \fcolorbox{black}{white}{\includegraphics[height=0.76in,width=0.95in]{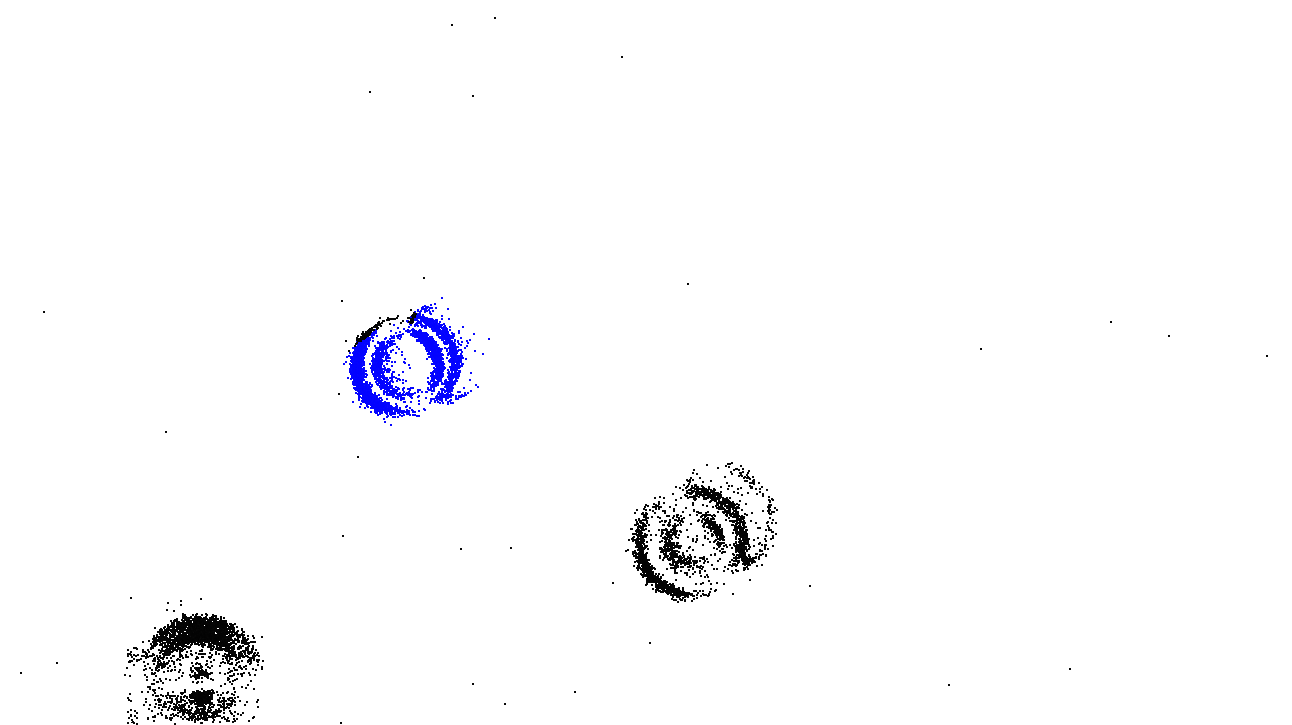}} \\

    \color{gray!90}Reference frame & \color{gray!90}T=7.4s (\color{red!90}Red) & \color{gray!90}T=10.2s (\color{red!90}Red) & \color{gray!90}T=23.8s (\color{green!90}Green) & \color{gray!90}T=25.4s (\color{green!90}Green) & \color{gray!90}T=41s (\color{blue!90}Blue) & \color{gray!90}T=44.5s (\color{blue!90}Blue) \\
    \fcolorbox{black}{white}{\includegraphics[height=0.76in,width=0.95in]{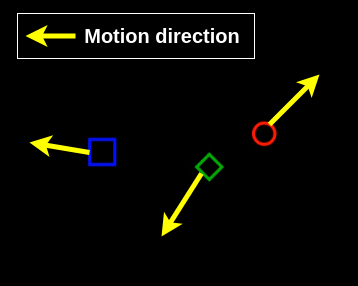}} &
    \fcolorbox{black}{white}{\includegraphics[height=0.76in,width=0.95in]{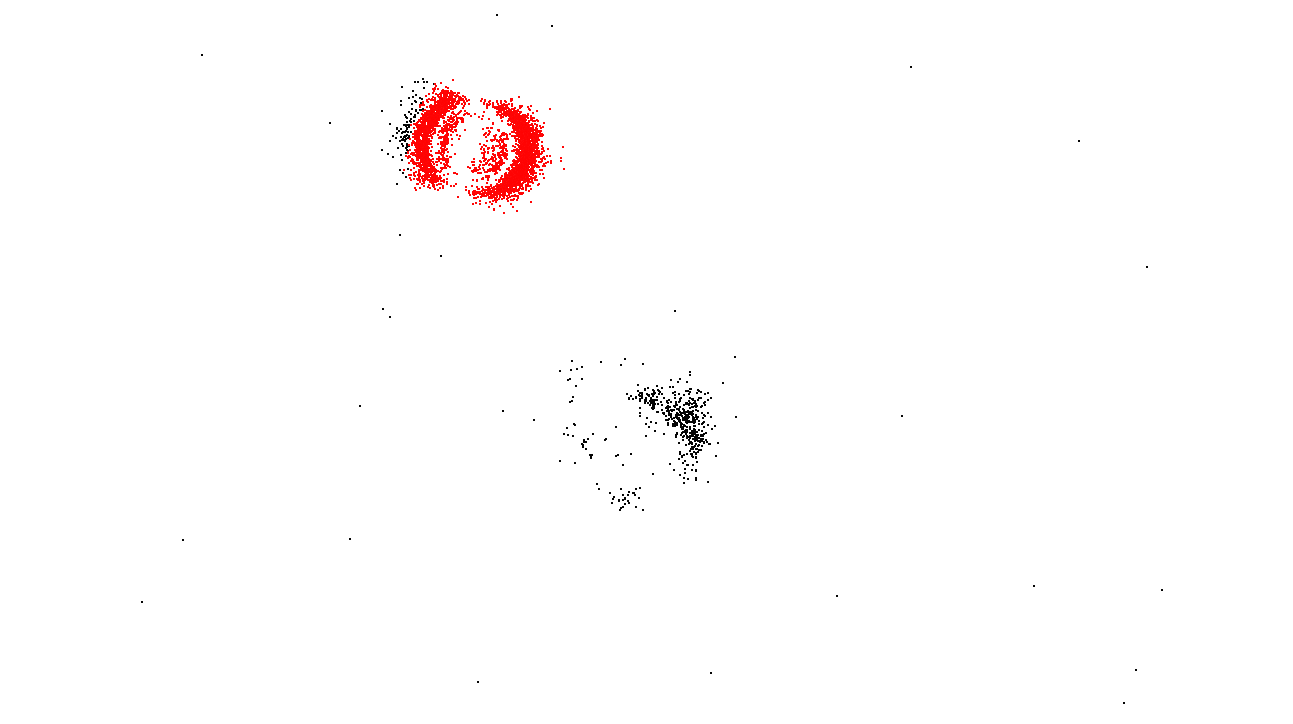}} &
    \fcolorbox{black}{white}{\includegraphics[height=0.76in,width=0.95in]{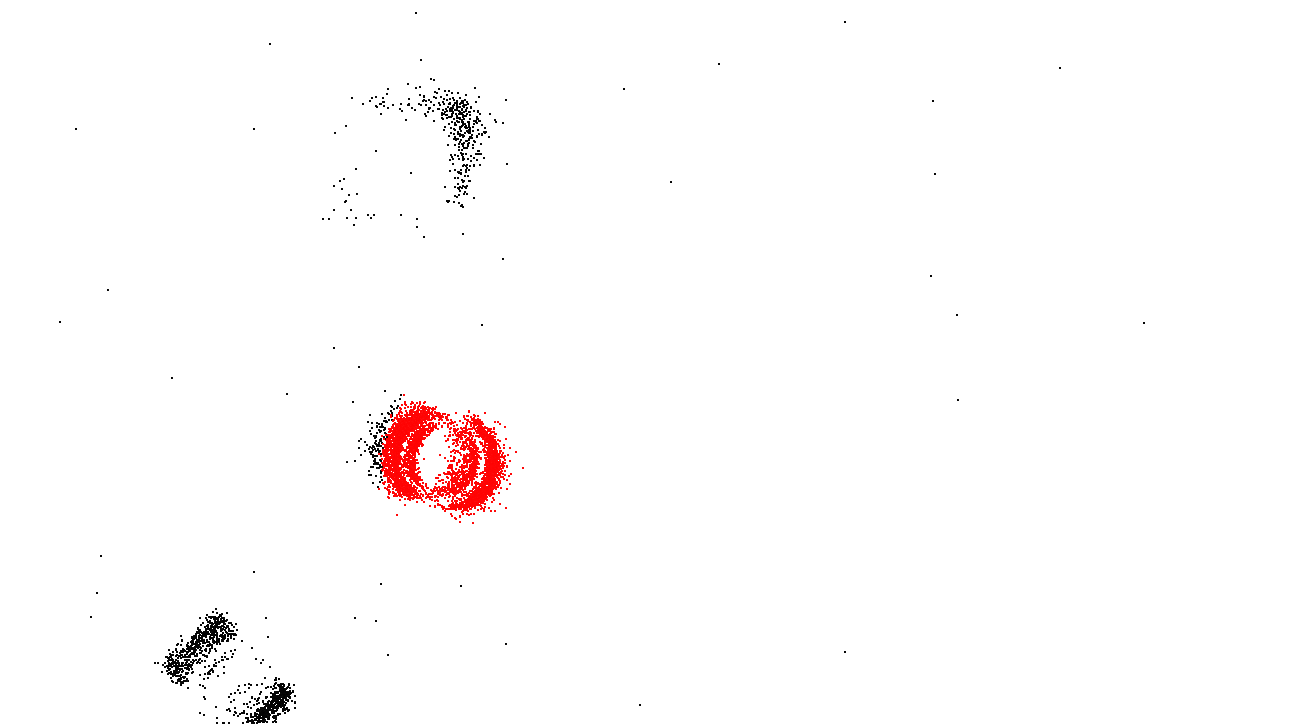}} &
    \fcolorbox{black}{white}{\includegraphics[height=0.76in,width=0.95in]{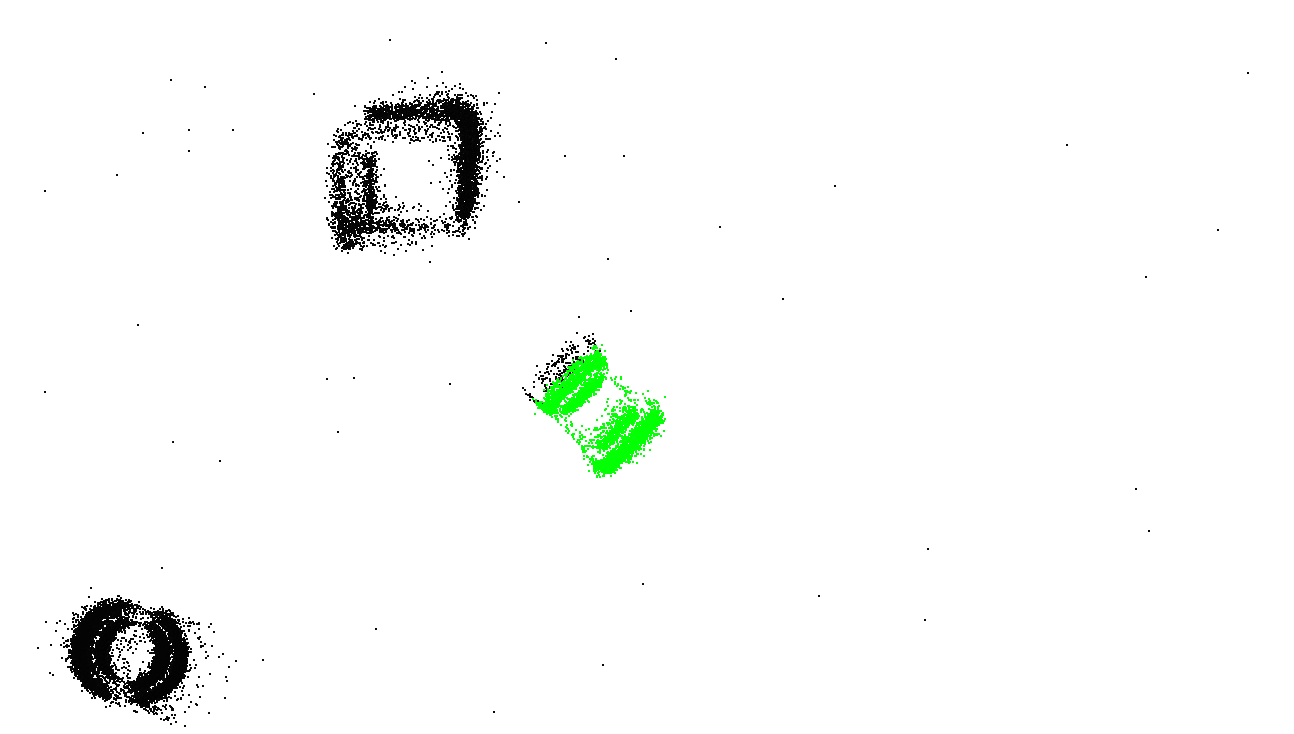}} &
    \fcolorbox{black}{white}{\includegraphics[height=0.76in,width=0.95in]{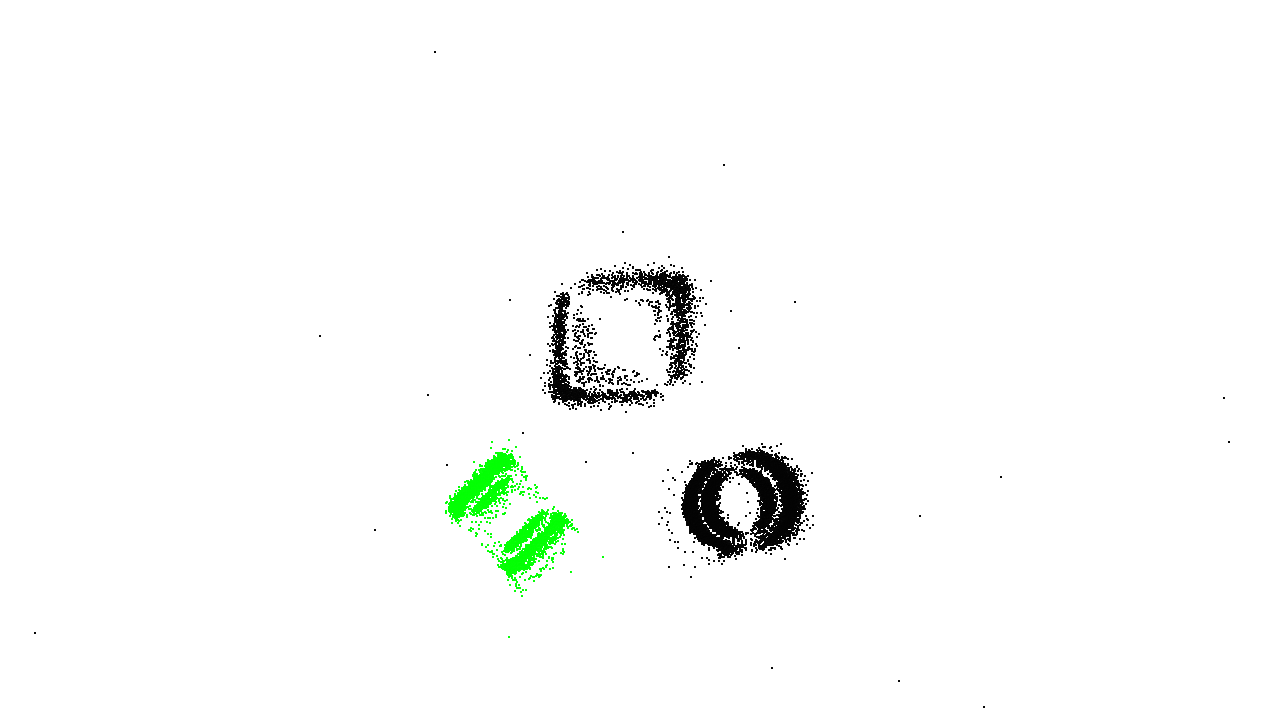}} &
    \fcolorbox{black}{white}{\includegraphics[height=0.76in,width=0.95in]{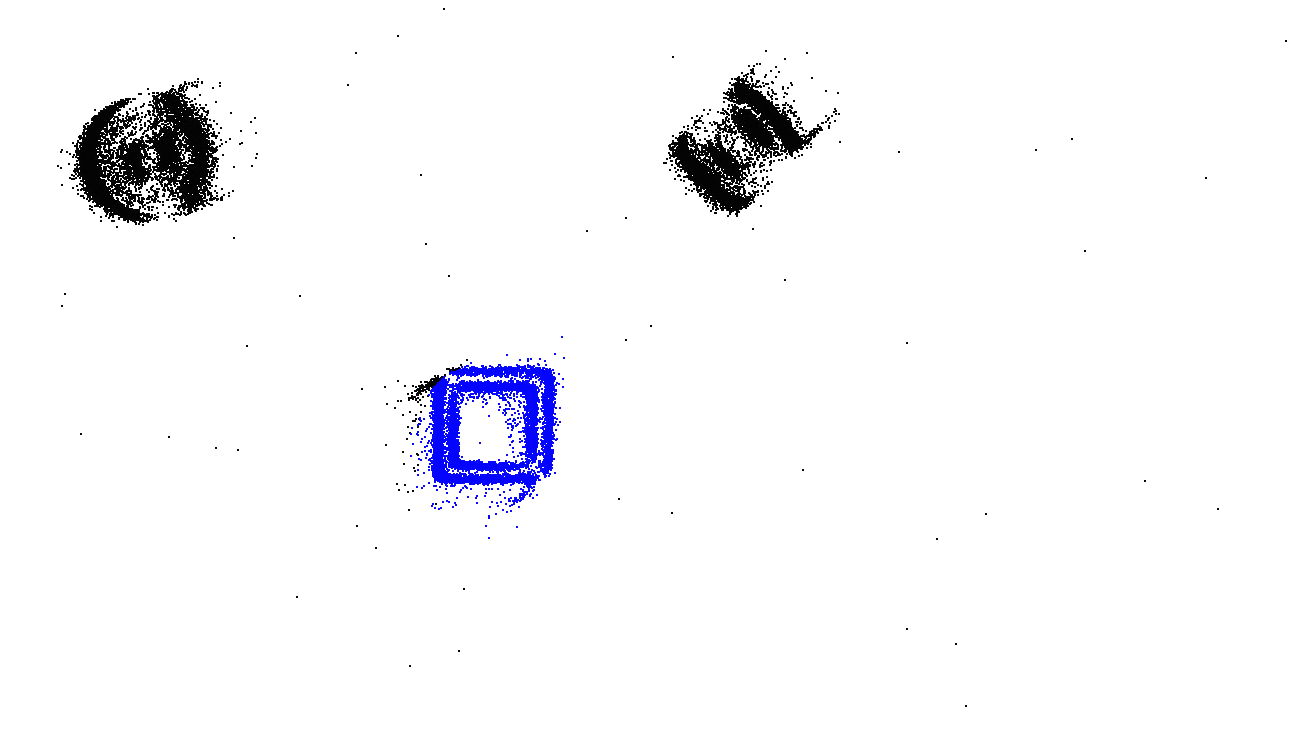}} &
    \fcolorbox{black}{white}{\includegraphics[height=0.76in,width=0.95in]{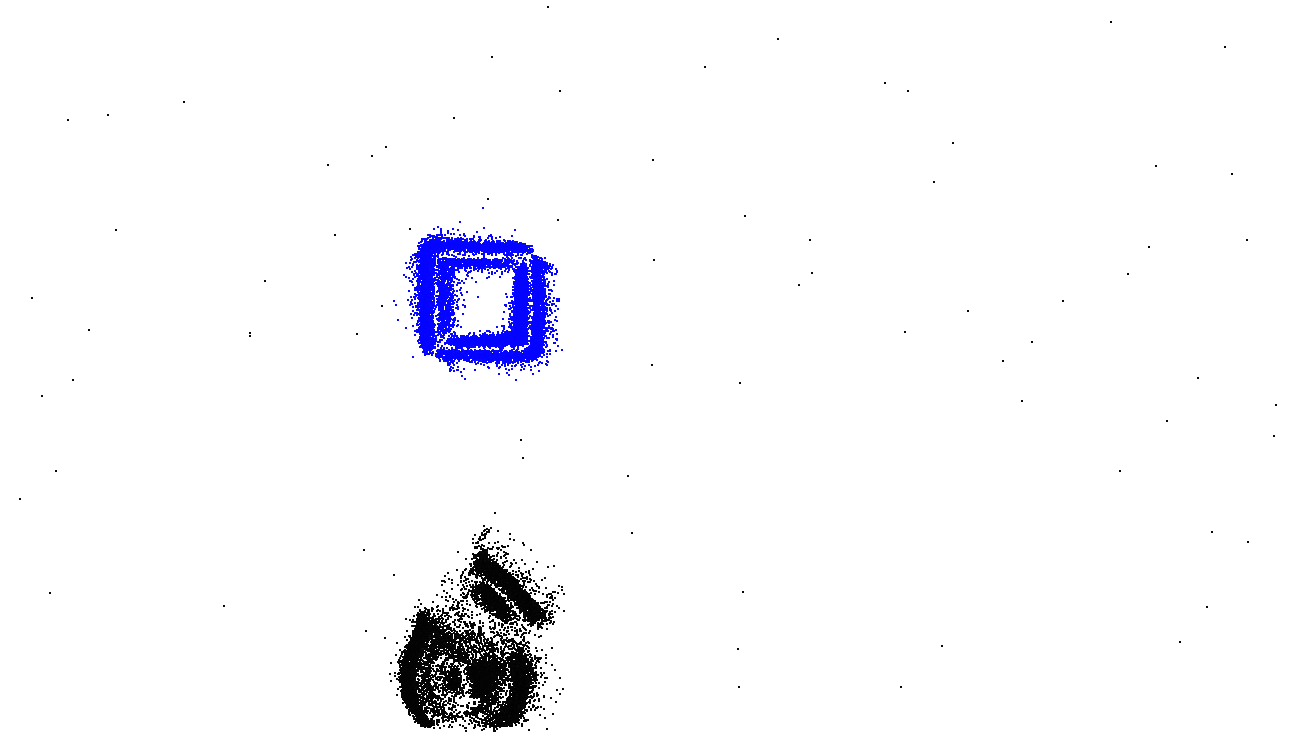}} \\
\end{tabular}
\caption{Colour segmentation on coloured moving objects. Sharp and blurry objects trigger events, but the sharp produce more contrast and therefore the algorithm can segment them at each focal plane.}
\label{tb:coloursegmentationvideo}
\end{figure*}

We evaluated this approach on both static (only focal length is changing) and dynamic (focal length changing with multiple moving objects) scenes, shown in Figures~\ref{fig:Coloursegmentation2d} and \ref{tb:coloursegmentationvideo}. Our results confirm the viability of colour-from-focus in uncontrolled settings. Figure~\ref{fig:Coloursegmentation2d}(a) illustrates how the camera-linear actuator setup can focus on one colour while blurring others, even when the displayed colours are not purely red, green, or blue. Although some overlap occurs (e.g., green events appearing in blurred red regions), the dominant signal remains sufficiently distinct for reliable detection. Adjusting the camera bias threshold or using a more precise actuator can further reduce overlap; this is left as future work. Figure~\ref{fig:Coloursegmentation2d}(b) shows how the focused images at different wavelengths can be combined to form a superimposed RGB image.

Finally, Figure~\ref{tb:coloursegmentationvideo} demonstrates successful segmentation for dynamic scenes with randomly moving shapes. In these cases, events are not discarded after each focal plane step, since object and camera motion cause event distributions to shift over time. Our results confirm that accurate spectral segmentation is achievable in moving scenarios, provided the linear actuator can reliably adjust focal positions.

\bibliographystyle{splncs04}
\bibliography{bibtex}